\pdfoutput=1

\documentclass[11pt]{article}

\usepackage[preprint]{acl}
\usepackage{anyfontsize}

\usepackage{times}
\usepackage{latexsym}

\usepackage[T1]{fontenc}

\usepackage[utf8]{inputenc}

\usepackage{microtype}

\usepackage{inconsolata}

\usepackage{graphicx}

\usepackage{amsmath}
\usepackage{multirow}
\usepackage{color, colortbl}
\usepackage{amssymb}
\usepackage{pifont}
\usepackage[export]{adjustbox}
\usepackage{subcaption}
\usepackage{algpseudocode}
\usepackage{algorithm}
\usepackage{booktabs}
\usepackage{hyperref}
\usepackage{makecell}
\usepackage{enumitem}
\usepackage{arydshln}
\usepackage{CJKutf8}

\newcommand{\xmark}{\ding{55}}%

\title{An Efficient Gloss-Free Sign Language Translation Using Spatial Configurations and Motion Dynamics with LLMs}

\author{
 \textbf{Eui Jun Hwang}\quad
 \textbf{Sukmin Cho}\quad
 \textbf{Junmyeong Lee}\quad
 \textbf{Jong C. Park\thanks{\scriptsize\texttt{Corresponding author.}}}
\\
 School of Computing \\
 Korea Advanced Institute of Science and Technology \\
 \texttt{\{ehwa20,nelllpic,david516,jongpark\}@kaist.ac.kr}
}

\begin{document}
\maketitle
\begin{abstract}
Gloss-free Sign Language Translation (SLT) converts sign videos into spoken language sentences without relying on glosses, which are the written representations of signs.
Recently, Large Language Models (LLMs) have shown remarkable translation performance in gloss-free methods by harnessing their powerful natural language generation capabilities. However, these methods often rely on domain-specific fine-tuning of visual encoders to achieve optimal results. By contrast, we emphasize the importance of capturing the spatial configurations and motion dynamics in sign language. With this in mind, we introduce \textbf{Spa}tial and \textbf{Mo}tion-based Sign Language Translation (\textbf{SpaMo}), a novel LLM-based SLT framework. 
The core idea of SpaMo is simple yet effective: instead of domain-specific tuning, we use off-the-shelf visual encoders to extract spatial and motion features, which are then input into an LLM along with a language prompt. Additionally, we employ a visual-text alignment process as a lightweight warm-up step before applying SLT supervision.
Our experiments demonstrate that SpaMo achieves state-of-the-art performance on three popular datasets---PHOENIX14T, CSL-Daily, and How2Sign---without visual fine-tuning\footnote{\scriptsize\texttt{Code is available at \url{https://github.com/eddie-euijun-hwang/SpaMo}}}.

\end{abstract}

\section{Introduction}

Sign language is a visual means of communication primarily used by Deaf communities, relying on physical movements rather than spoken words. In this paper, we tackle Sign Language Translation (SLT), focusing on converting sign videos into spoken language sentences. 
Early SLT methods~\cite{camgoz2020sign,zhou2021improving,chen2022simple,chen2022two,zhang2023sltunet} have primarily relied on \textit{glosses}---written representations of signs using corresponding words. Glosses provide a structured form of sign language, which helps identify semantic boundaries within continuous sign sequences. This, in turn, allows the models to better comprehend the overall content of the sign videos~\cite{yin2023gloss,wei2023improving}. 
However, annotating glosses is a labor-intensive and time-consuming process that requires expertise in sign language. This significantly hinders the expansion of sign language datasets and limits the development of SLT methods~\cite{li2020tspnet,shi2022open,lin2023gloss}.

\begin{figure}[t]
    \begin{subfigure}{\linewidth}
    \centering
        \includegraphics[trim=0cm 1.5cm 0cm 0cm,clip=true,width=0.9\textwidth]{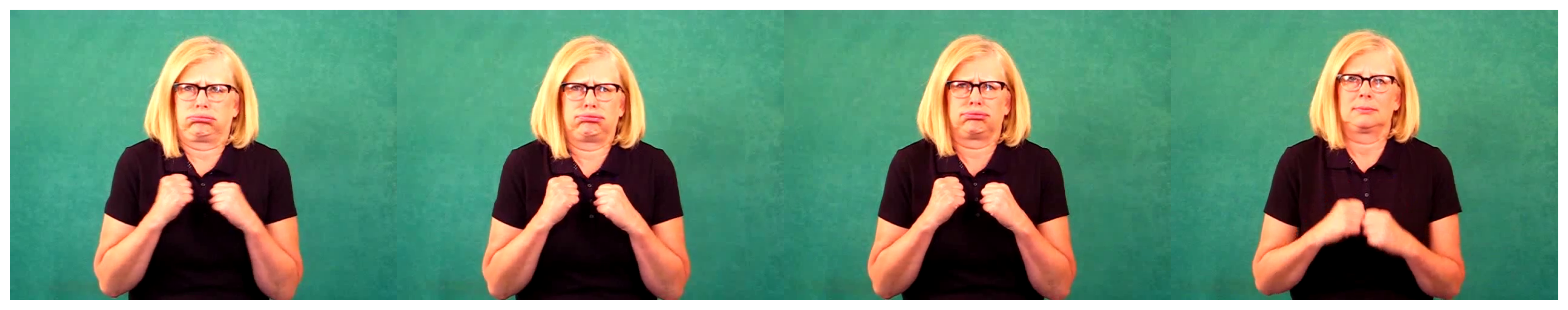}
        \vfill
        \small\text{``Cold''}
        \vfill
        \includegraphics[trim=0cm 1.5cm 0cm 0cm,clip=true,width=0.9\textwidth]{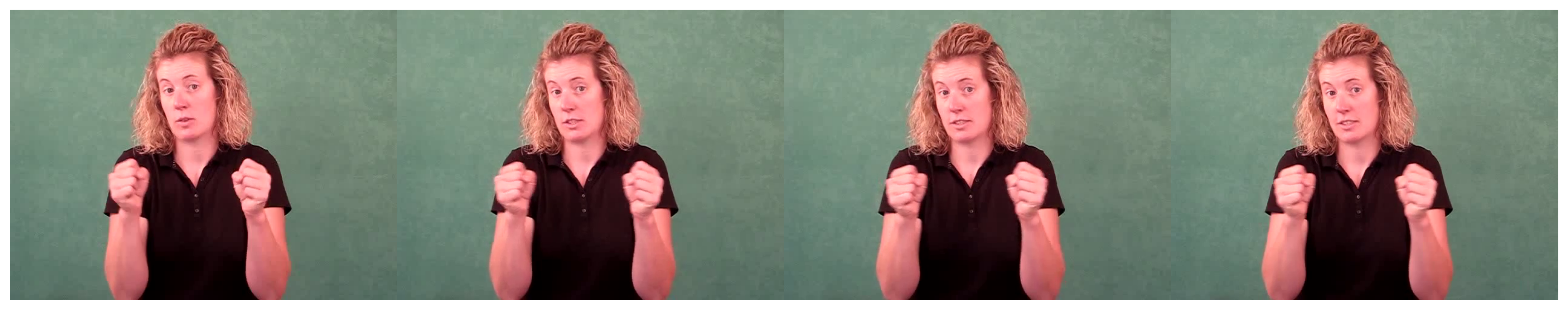}
        \vfill
        \small\text{``Winter''}
    \caption{Spatial configuration} 
    \label{fig:cold_winter}
    \end{subfigure}
    \vfill \vspace{0.3em}
    \begin{subfigure}{\linewidth}
    \centering
        \includegraphics[trim=0cm 0cm 0cm 0cm,clip=true,width=0.9\textwidth]{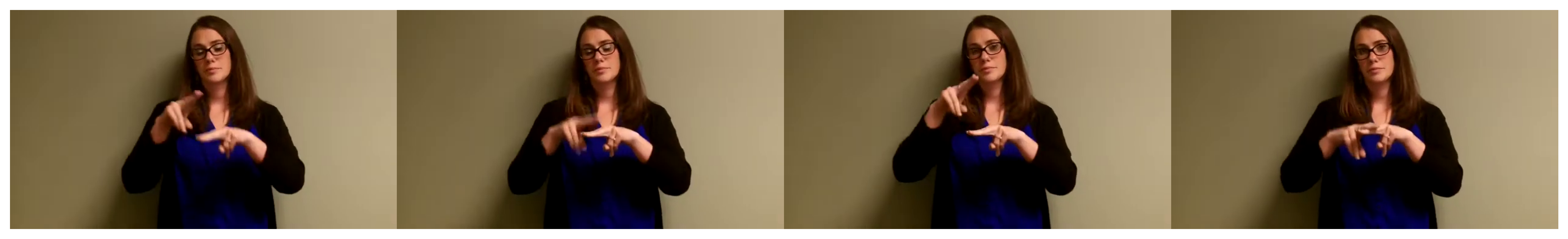}
        \vfill
        \small\text{``Chair''}
        \vfill
        \includegraphics[trim=0cm 0cm 0cm 0cm,clip=true,width=0.9\textwidth]{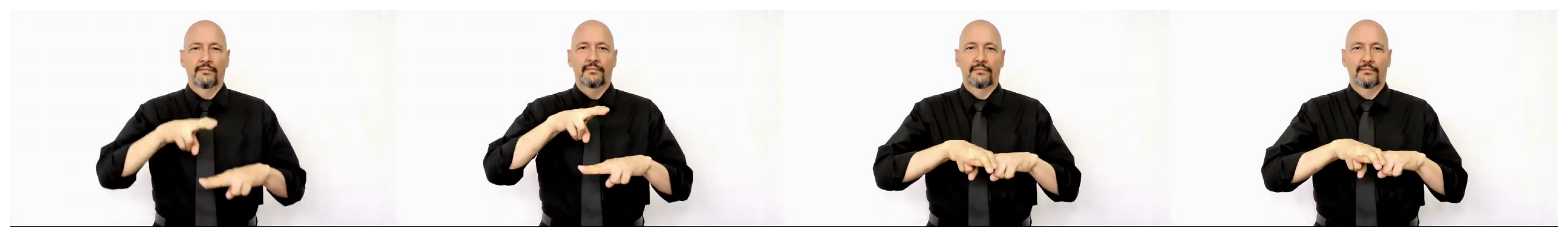}
        \vfill
        \small\text{``Sit''}
    \caption{Motion dynamic} 
    \label{fig:chair_sit}
    \end{subfigure}
    \vspace{-2em}
\caption{Visual examples of spatial configurations and motion dynamics in sign language. The images are sourced from WLASL~\cite{li2020word}.}
\label{fig:sc_md_vis_example}
\vspace{-1.5em}
\end{figure}
   
To address these limitations, there has been a shift towards gloss-free methods that rely solely on the sign videos and corresponding translations. While these methods still underperform compared to the gloss-based methods, efforts have been made to reduce the performance gap by focusing on temporal semantic structures~\cite{li2020tspnet} and aligning visual and textual modalities~\cite{zhao2021conditional,yin2023gloss,zhou2023gloss,zhao2024conditional}. 
Recently, LLMs have demonstrated remarkable translation performance in a gloss-free setting by harnessing their powerful language generation capabilities. 
However, the modality gap between the continuous sign videos and discrete text poses a challenge for the LLMs in effectively understanding the sign videos. 
To address this, many methods fine-tune their visual encoders to be more domain-specific to sign language~\cite{wong2024sign2gpt,chen2024factorized,rust2024towards,gong2024llms}.

That said, fine-tuning visual encoders can be resource-heavy and time-consuming, making it impractical for real-world applications, especially when considering the diversity of sign languages.
This leads to an important question: \textit{Is domain-specific tuning of visual encoders necessary to achieve optimal performance in LLM-based SLT?}
We argue that focusing on the inherent characteristics of sign language could reduce the need for such extensive fine-tuning. 
First, visual encoders trained on general domains~\cite{radford2021learning,oquab2023dinov2} have already proven highly effective in downstream tasks such as action recognition~\cite{huang2024froster,tang2024survey} and video captioning~\cite{yang2023vid2seq,zhou2024streaming}. 
Moreover, LLMs are capable of retaining rich visual information from these general encoders in their latent space~\cite{zhang2024visually}.
Rather than emphasizing fine-tuning, we shift our attention to the crucial roles of \textbf{spatial configurations} and \textbf{motion dynamics} in sign language.

Spatial configurations include the arranging and positioning of signs within the signing space, including hand shapes, facial expressions, and body postures. 
These components work together to distinguish different signs and convey their intended meanings~\cite{emmorey1995comparison}. 
As shown in Figure~\ref{fig:cold_winter}, the signs for ``cold'' and ``winter'' both use the same handshape, with a shivering motion of the fists. The primary difference lies in the facial expressions: ``cold'' is typically accompanied by a tense or grimaced expression, while ``winter'' may feature a more neutral expression. 
Motion dynamics, on the other hand, involve the path, speed, and rhythm of hand movements, illustrating how movements alter the meanings of signs over time~\cite{bosworth2019analysis}. 
As shown in Figure~\ref{fig:chair_sit}, the signs for ``chair'' and ``sit'' both use the same ``H'' handshape and involve the interaction of both hands. However, the motion differentiates these signs: ``chair'' involves a repetitive tapping motion, while ``sit'' involves a single, smooth motion. These examples highlight the importance of the spatial configurations and motion dynamics in conveying accurate messages in sign language.

To this end, we introduce a novel gloss-free framework, \textbf{Spa}tial and \textbf{Mo}tion-based Sign Language Translation (\textbf{SpaMo}). 
SpaMo is designed to fully exploit the spatial configurations and motion dynamics in the sign videos using off-the-shelf visual encoders, all without the need for domain-specific fine-tuning. 
As shown in Figure~\ref{fig:method_overview}, the core idea is simple: We extract spatial features (spatial configurations) and motion features (motion dynamics) using two different visual encoders, and feed these into an LLM with a language prompt. 
Specifically, we use a pre-trained image encoder (e.g., ViT) as \textbf{Spatial Encoder (SE)} to individually encode each frame for its spatial features. To further refine the spatial features, we apply \(S^2\) scaling~\cite{shi2024we}, which processes a sign image at multiple scales. Additionally, we use a video encoder (e.g., VideoMAE) as \textbf{Motion Encoder (ME)} to encode sign clips (groups of sign frames) into the motion features. To capture finer motion dynamics, we apply a sliding window approach, which results in implicit gloss-level representations~\cite{cheng2023cico,hwang2024universal}.
Next, \textbf{Sign Adapter (SA)}, comprising Multi-Layer Perceptron (MLP) layers, transfers these features to the LLM. 
To further bridge the modality gap, we propose \textbf{Visual-Text Alignment (VT-Align)}, a training strategy that aligns the visual features with the LLM's embedding space, promoting more efficient training and improved translation performance.

In all, our contributions can be summarized as:
\vspace{-0.1in}
\begin{itemize}[itemsep=0.3mm, parsep=1pt, leftmargin=*]
    \item We introduce SPaMo, a novel gloss-free framework based on LLMs. Our method is simple yet effective, focusing on conveying core elements of sign language to LLMs without domain-specific tuning of visual encoders.
    \item Our proposed method achieves state-of-the-art performance on three popular sign language datasets: PHOENIX14T, CSL-Daily, and How2Sign.
    \item We provide a novel and comprehensive analysis of how the LLM interprets the sign videos within its embedding space and translates them into corresponding text.
\end{itemize}

\section{Related Work}\label{sec:related_work}

\subsection{Gloss-free Sign Language Translation}

Gloss-free SLT directly converts sign videos into spoken language sentences without relying on glosses. These methods, however, often underperform compared to gloss-based methods~\cite{camgoz2020sign,zhou2021spatial,zhou2021improving,yin2021simulslt,chen2022simple,chen2022two,zhang2023sltunet,jing2024vk}. To address the performance gap, recent work has focused on several key areas: enhancing the temporal semantic structure~\cite{li2020tspnet}, improving the alignment between visual and textual modalities~\cite{zhao2021conditional,lin2023gloss,fu2023token}, leveraging LLMs~\cite{wong2024sign2gpt,gong2024llms,chen2024factorized}, and scaling efforts by utilizing larger sign language datasets~\cite{uthus2024youtube,rust2024towards}.
Despite these advancements, most gloss-free methods depend on fine-tuning visual encoders using the glosses~\cite{li2020tspnet,yin2023gloss,fu2023token}, target translations~\cite{zhou2023gloss,wong2024sign2gpt}, or self-supervised learning~\cite{gong2024llms,rust2024towards}. In particular, fine-tuning with the glosses helps the visual encoders to offer more domain-specific training on continuous or isolated Sign Language Recognition (SLR) datasets, such as WLASL~\cite{li2020word} and PHOENIX14T~\cite{camgoz2018neural}. 

Consequently, we classify these methods as \textit{weakly gloss-free} due to the implicit involvement of the glosses, as further elaborated in Section~\ref{sec:exp_settings}. On the other hand, the rest of the fine-tuning methods eliminate reliance on these annotations. 
However, they often require substantial resources. As a results, it can be difficult to achieve robust visual representations and improve translation performance without access to a sufficiently large dataset. 
To address this limitation, our approach diverges from this norm by focusing on capturing the spatial configurations and motion dynamics through off-the-shelf visual encoders. This allows us to bypass the need for resource-intensive fine-tuning.

\subsection{Scaling Language Models in SLT}

The scaling laws in language models~\cite{Kaplan2020ScalingLF} have been pivotal in the rise of Large Language Models (LLMs)~\cite{touvron2023llama,chiang2023vicuna,chung2024scaling}. 
Leveraging their strong generation capabilities, LLMs have been applied across diverse domains: multilingual translation~\cite{zhu2023multilingual,zhang2023prompting,gao2024towards}, pose generation~\cite{feng2024chatpose,zhang2024motiongpt}, and visual question answering~\cite{li2023blip,liu2024improved,liu2024visual}, extending their impact beyond Natural Language Processing.
LLMs have also demonstrated impressive translation performance in the SLT domain. SLT methods using the LLMs focus on aligning high-dimensional visual features with inputs comprehensible to LLMs. 
These methods involve fine-tuning visual encoders to produce language-like tokens~\cite{gong2024llms}, using pseudo-glosses~\cite{wong2024sign2gpt}, or performing video-grounded text generation tasks~\cite{chen2024factorized}.

In this work, we take a different approach by focusing on spatial configurations and motion dynamics. We extract spatial and motion features and pass them to LLMs with a light warm-up process. This method is simple yet effective, demonstrating that an extensive pre-training for the visual encoders is unnecessary to achieve peak performance.

\begin{figure*}[t]
    \centering
    \includegraphics[width=\textwidth]{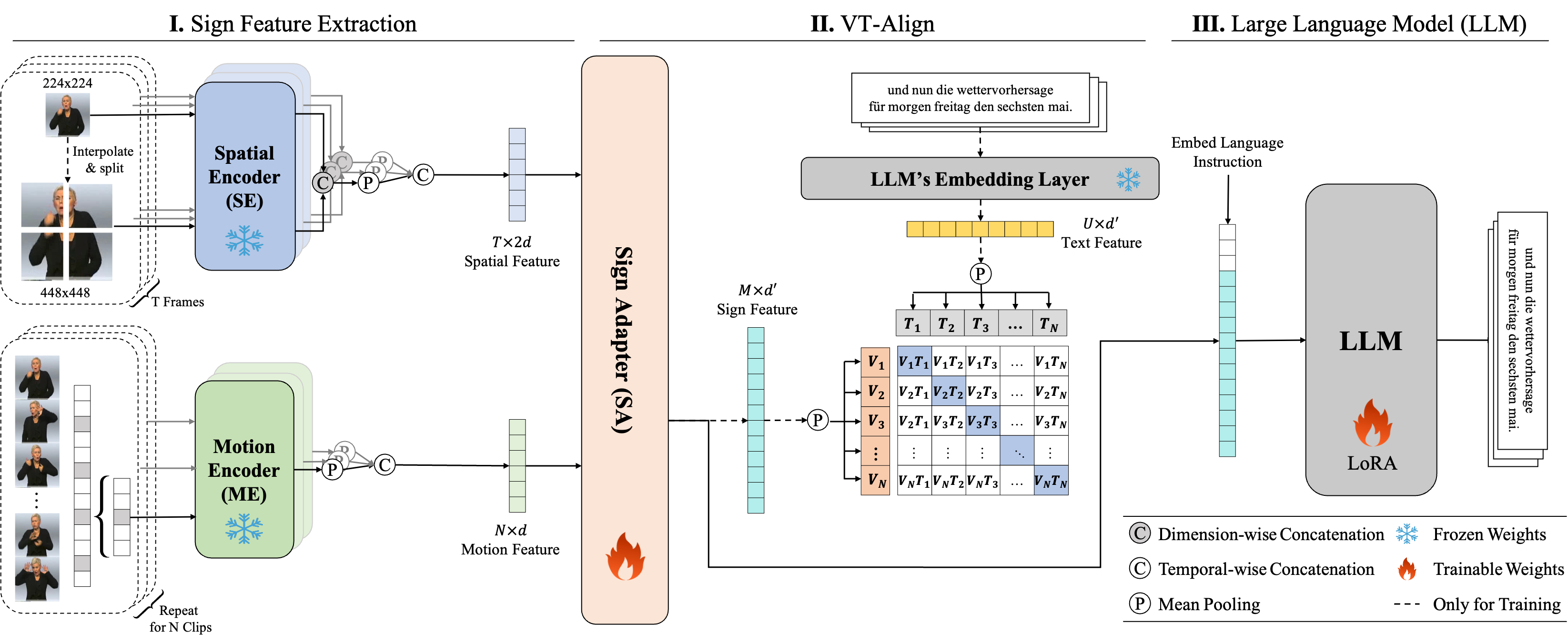}
    \vspace{-1.5em}
    \caption{An overview of the SpaMo framework, which consists of three parts: (i) \textbf{Sign Feature Extraction}: Spatial and motion features are extracted using SE and ME, using the \(S^2\) and sliding window approaches to capture detailed spatial configurations and motion dynamics. (ii) \textbf{VT-Align}: The extracted features are combined within SA to form a unified sign feature. During training, a warm-up process is employed to ensure that SA has well-initialized weights, effectively bridging the modality gap between the sign video and text. (iii) \textbf{LLM}: the LLM processes the sign feature along with a language-instructive prompt and is trained using LoRA.} 
    \label{fig:method_overview}
\vspace{-1em}
\end{figure*}

\section{Method}

We first give an overview of our framework in Section~\ref{sec:overview}. We then explain Spatial Encoder (SE) and Motion Encoder (ME) in Sections~\ref{sec:spatial_enc} and~\ref{sec:motion_enc}, respectively. Next, we discuss Sign Adapter (SA) in Section~\ref{sec:sa} and VT-Align in Section~\ref{sec:vt_align}. Finally, we explain the training details in Section~\ref{sec:tr_infer}.

\subsection{Framework Overview}\label{sec:overview}

Given a sign video \(X=\{x_i\}_{i=1}^T\), where each frame \(x_i \in \mathbb{R}^{H \times W}\) represents a frame with height \(H\) and width \(W\), the objective of SLT is to generate a corresponding translation \(Y=\{y_j\}_{j=1}^U\), composed of \(U\) words. 
Previous gloss-free methods~\cite{zhou2023gloss,wong2024sign2gpt,gong2024llms,chen2024factorized} have involved fine-tuning visual encoders using sign language data to make them more domain-specific, leading to improvements in translation accuracy. 
However, while this fine-tuning introduces more domain knowledge at the feature extraction level, it is unnecessary, especially with LLMs, which already maintain rich visual information from the visual encoder in their latent space~\cite{zhang2024visually}. 
Although there may be a trade-off, we argue that utilizing the spatial configurations and motion dynamics with proper alignment and training, offers a more efficient and effective solution.

As shown in Figure~\ref{fig:method_overview}, SE and ME extract two distinct features from the sign video \(X\): Spatial features \(Z_s\) capture the spatial configurations~\cite{emmorey1995comparison}, and motion features \(Z_m\) represent the motion dynamics~\cite{bosworth2019analysis}. These features are then integrated into a combined sign feature \(Z_{sm}\) via SA. The combined feature is then fed to an LLM with a language prompt, guiding the LLM to generate the translation in the desired language. Additionally, we perform Visual-Text Alignment (VT-Align) to minimize the gap between the visual and textual modalities before and during training under SLT supervision. 
In the following sections, we provide a detailed explanation of SE, ME, SA, and VT-Align. 

\subsection{Spatial Encoder}\label{sec:spatial_enc}

SE extracts spatial features \(Z_s\) from the sign video \(X\). We utilize a pre-trained image encoder (e.g., ViT), which is kept frozen, and enhances its capability to capture finer spatial information by applying Scaling on Scales (\(S^2\))~\cite{shi2024we}. \(S^2\) is parameter-free and enables the extraction of multi-scale features without altering the original pre-trained encoder. By processing sign images at multiple resolutions, \(S^2\) provides a more comprehensive spatial understanding, ensuring that SE captures both fine-grained and broad spatial details for accurate sign language interpretation. The resulting spatial features can be represented as \(Z_s \in \mathbb{R}^{T \times 2d}\), where \(T\) is the number of frames, and \(2d\) is the increased embedding dimension, reflecting the integration of multi-scale features.

\subsection{Motion Encoder}\label{sec:motion_enc}

ME derives motion features from the sign video \(X\). Similar to SE, we employ a pre-trained video encoder (e.g., VideoMAE), which remains frozen, to process sign clips segmented from the video. However, accurately segmenting the sign video into distinct gloss-level clips is challenging without the support of pre-trained Continuous Sign Language Recognition (CSLR) models~\cite{wei2023improving}. 
To address this limitation, we use a sliding window approach to capture implicit gloss-level representations~\cite{cheng2023cico,hwang2024universal}. 
Specifically, we divide the sign video into short, overlapping clips, then feed each clip into ME to extract the implicit gloss-level motion features \(Z_m \in \mathbb{R}^{N\times d}\), where \(N\) is the number of segments. The number of segments \(N\) is calculated as \(N = \left \lfloor \frac{T-w}{s} \right \rfloor + 1\), where \(T\) is the total number of frames, and \(w\) and \(s\) are the window size and stride, respectively. Since \(Z_m\) is generated by processing \(N\) short clips, it can also be interpreted as a sequence of \(N\) clip-wise features.

\begin{figure}[t]
    \centering
    \includegraphics[width=0.8\linewidth]{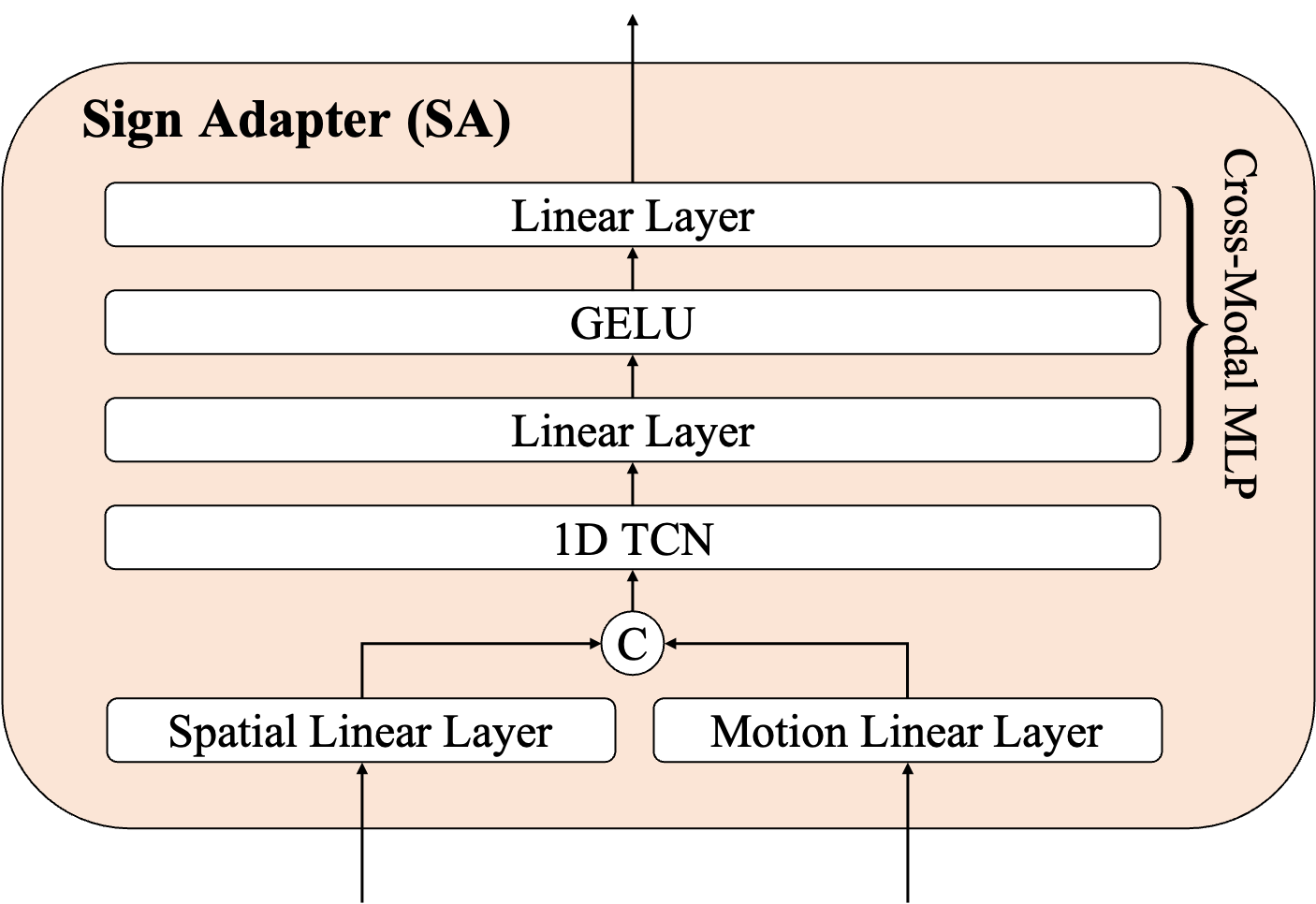}
    \vspace{-0.5em}
    \caption{An overview of Sign Adapter.} 
    \label{fig:sa_overview}
    \vspace{-1.5em}
\end{figure}

\subsection{Sign Adapter}\label{sec:sa}

In the previous sections, we extracted two distinct visual features: the spatial features \(Z_s\) and the motion features \(Z_m\). These features differ in both their dimensions and representation, as depicted in Figure~\ref{fig:method_overview}.
To effectively integrate these features, we introduce an additional module called Sign Adaptor (SA). As shown in Figure~\ref{fig:sa_overview}, SA includes linear projection layers, a 1D TCN, and a Multi-Layer Perceptron (MLP). These components work together to integrate the spatial and motion features into a unified sign representation, denoted as \(Z_{sm}\).
First, the spatial and motion features are passed through linear projection layers to transform them into features with matching dimensions. Next, the 1D TCN is applied for short-term modeling of the combined features. Finally, a cross-modal MLP~\cite{liu2024improved} is employed to bridge the visual and textual modalities. 
The resulting outputs are represented as \(Z_{sm} \in \mathbb{R}^{M \times d'}\), where \(M\) represents the reduced number of frames after convolution, and \(d'\) is the dimension aligned with that of the LLM. Although SA aids in bridging the modality gap between visual and textual features during training under the SLT supervision, the gap remains. To tackle this issue, we introduce VT-Align, which will be detailed in the next section.

\subsection{Visual-Text Alignment}\label{sec:vt_align}

VT-Align is a \textit{warm-up and go} process designed to provide the SA module with well-initialized weights before the SLT supervision begins. This initial alignment is crucial, as it helps the model more effectively bridge the modality gap during training. To achieve this alignment, we employ a widely-used softmax-based contrastive learning approach~\cite{radford2021learning,jia2021scaling}. 

Specifically, given a mini-batch \(\mathcal{B}=\{(S_1,Y_1),(S_2,Y_2),...\}\) of sign-text pairs, the contrastive learning objective encourages the embeddings of matching pairs \((S_i,Y_i)\) to align closely while pushing apart the embeddings of mismatched pairs \((S_i,Y_{j \neq i})\). 
Text features \(Z_t\) are extracted from the target translation \(Y_i\) using the LLM's embedding layer \(E_{llm}(\cdot)\). Note that only the SA module \(f_{sa}(\cdot)\) is updated during this process, while \(E_{llm}(\cdot)\) remains fixed to preserve the LLM's language capabilities. The VT-Align loss function \(\mathcal{L}_{vt}\) is represented as follows:
\begin{equation}
    \tiny
    -\frac{1}{2|\mathcal{B}|} \sum_{i=1}^{|\mathcal{B}|} \left( 
    \overbrace{\log \frac{e^{\tau{Z_{sm}^{(i)}}\cdot z_{t}^{(i)}}}{\sum_{j=1}^{|\mathcal{B}|} e^{\tau{Z_{sm}^{(i)} \cdot Z_{t}^{(j)}}}}}^{\text{sign}\rightarrow \text{text softmax}} +
    \overbrace{\log \frac{e^{\tau{Z_{sm}^{(i)}}\cdot Z_{t}^{(i)}}}{\sum_{j=1}^{|\mathcal{B}|} e^{\tau{Z_{sm}^{(j)} \cdot Z_{t}^{(i)}}}}}^{\text{text}\rightarrow\text{sign softmax}}
    \right),
\end{equation}
where \(Z_{sm}^{(i)}=\frac{f_{sa}(S_i)}{\left\| f_{sa}(S_i) \right\|_2}\), \(Z_{t}^{(i)}=\frac{E_{llm}(T_i)}{\left\| E_{llm}(T_i) \right\|_2}\), and \(\tau\) denotes a learnable temperature parameter used to scale the logits.

\subsection{Training Details}\label{sec:tr_infer}

Our framework is optimized in two phases: an initial warm-up phase followed by training with the SLT supervision. 
In the warm-up phase, we begin with a training phase the SA module using VT-Align for a designated number of steps (e.g., 4K steps). 
After completing the warm-up phase, we proceed to the joint training for both SA and the LLM. 
For fine-tuning the LLM, we utilize LoRA~\cite{hu2021lora}, a lightweight and efficient method specifically designed for this purpose. 
Overall, our method is trained with a combined loss function:
\begin{equation}    \mathcal{L}_\textit{SpaMo}=\mathcal{L}_{ce}+\mathcal{L}_{vt},
\end{equation}
where \(\mathcal{L}_{ce}\) represents cross-entropy loss, and \(\mathcal{L}_{vt}\) continuously manages the alignment. 


\begin{table*}
    \scriptsize
    \centering
    \resizebox{0.95\linewidth}{!}{%
        \begin{tabular}{c l c ccccc c ccccc} \toprule
         && \multicolumn{5}{c}{\textbf{PHOENIX14T}} & \multicolumn{5}{c}{\textbf{CSL-Daily}} \\
        \cmidrule(lr){4-8} \cmidrule(lr){9-13}
        \textbf{Setting} & \textbf{Methods} & \textbf{Vis. Ft.} & \textbf{B1} & \textbf{B2} & \textbf{B3} & \textbf{B4} & \textbf{RG} & \textbf{B1} & \textbf{B2} & \textbf{B3} & \textbf{B4} & \textbf{RG} \\
        \midrule
        \multirow{5}{*}{Gloss-based} 
        & SLRT~\cite{camgoz2020sign}& \(\checkmark\)& 46.61& 33.73& 26.19& 21.32& - & 37.38& 24.36& 16.55& 11.79& 36.74 \\
        & BN-TIN-Transf.+SignBT~\cite{zhou2021improving}& \xmark& 50.80 &37.75 &29.72 &24.32 &49.54 &51.42 &37.26 &27.76 &21.34 &49.31 \\
        & MMTLB~\cite{chen2022simple} &\(\checkmark\) &53.97 &41.75 &33.84 &28.39 &52.65 &53.31& 40.41 &30.87 &23.92 &53.25 \\
        & TS-SLT~\cite{chen2022two} &\(\checkmark\) &54.90 &42.43 &34.46 &28.95 &53.48 &55.44& 42.59 &32.87 &25.79 &55.72 \\
        & SLTUNET~\cite{zhang2023sltunet} &\(\checkmark\) &52.92 &41.76 &33.99 &28.47 &52.11 &54.98 &41.44 &31.84 &25.01 &54.08 \\ \midrule
        \multirow{3}{*}{Weakly Gloss-free} 
        & TSPNet~\cite{li2020tspnet}\(\ddagger\)& \(\checkmark\) & 36.10& 23.12& 16.88& 13.41&34.96 & 17.09& 8.98& 5.07& 2.97& 18.38 \\
        & GASLT~\cite{yin2023gloss}& \(\checkmark\)& 39.07& 26.74& 21.86& 15.74&39.86 & 19.90& 9.94& 5.98& 4.07& 20.35\\ 
        & ConSLT~\cite{fu2023token}&\(\checkmark\) &- &- &- &21.59 &47.69 &- &- &- &- &- \\ \midrule
        \multirow{6}{*}{Gloss-free} 
        & CSGCR~\cite{zhao2021conditional}& \xmark& 36.71& 25.40& 18.86& 15.18&38.85 &- &- &- &- &- \\
        & GFSLT-VLP~\cite{zhou2023gloss} &\(\checkmark\) &43.71 &33.18 &26.11 &21.44 &42.29 &39.37 &24.93 &16.26 &11.00 &36.44\\
        & FLa-LLM~\cite{chen2024factorized}& \(\checkmark\)& 46.29& 35.33& 28.03& 23.09& 45.27 &37.13& 25.12& 18.38& 14.20& 37.25 \\
        & Sign2GPT~\cite{wong2024sign2gpt}& \(\checkmark\)& \underline{49.54} &\underline{35.96}& \underline{28.83}& 22.52& \textbf{48.90} &\underline{41.75}& \underline{28.73}& \underline{20.60}& 15.40& \underline{42.36} \\ 
        & SignLLM~\cite{gong2024llms} &\(\checkmark\) &45.21 & 34.78 & 28.05 &\underline{23.40} &44.49 &39.55 & 28.13 & 20.07 &\underline{15.75} &39.91 \\ 
        \noalign{\vskip 0.3ex}\cdashline{2-14}\noalign{\vskip 0.7ex}
        & \textbf{SpaMo (Ours)}& \xmark& \textbf{49.80}& \textbf{37.32} &\textbf{29.50} &\textbf{24.32} &\underline{46.57} & \textbf{48.90}& \textbf{36.90}& \textbf{26.78}& \textbf{20.55}& \textbf{47.46} \\ \bottomrule
        \end{tabular}
    }
    \caption{Performance comparison on the PHOENIX14T and CSL-Daily datasets. ``Vis. Ft.'' denotes the visually fine-tuned on sign language datasets. \(\ddagger\) denotes results reproduced by~\citeauthor{yin2023gloss} for CSL-Daily. The best results are highlighted as \textbf{bold}, and the second-best are \underline{underlined}.}
    \label{tab:res_p14t_csl}
    \vspace{-1em}
\end{table*}
\begin{table*}
\scriptsize
\centering
\resizebox{0.95\linewidth}{!}{%
    \begin{tabular}{c l c c cccccc} \toprule
    \textbf{Setting}& \textbf{Methods} &\textbf{Modality}& \textbf{Vis. Ft.}& \textbf{BLEU-1} &\textbf{BLEU-2} &\textbf{BLEU-3} &\textbf{BLEU-4} &\textbf{ROUGE} &\textbf{BLEURT}\\ \midrule
    \multirow{2}{*}{Weakly Gloss-free}& GloFE-VN~\cite{lin2023gloss}& Landmark& \(\checkmark\)& 14.94 &7.27 &3.93 &2.24 &12.61 &31.65 \\
                                        & OpenSLT~\cite{tarres2023sign}& RGB& \(\checkmark\)& 34.01 &19.30 &12.18 &8.03 &- &- \\ \midrule
    \multirow{4}{*}{Gloss-free}& YT-ASL-SLT~\cite{uthus2024youtube}\(\dagger\)& Landmark& \xmark& 14.96 & 5.11& 2.26& 1.22& -& 29.98 \\ 
                                & SSVP-SLT~\cite{rust2024towards}\(\dagger\)& RGB& \(\checkmark\)& \underline{30.20}& 16.70&  10.50& 7.00& 25.70& \underline{39.30} \\
                                & FLa-LLM~\cite{chen2024factorized}& RGB& \(\checkmark\)& 29.81& \underline{18.99}& \underline{13.27}& \underline{9.66}& \underline{27.81}& - \\ \noalign{\vskip 0.3ex}\cdashline{2-10}\noalign{\vskip 0.7ex}
                                & \textbf{SpaMo (Ours)} & RGB& \xmark& \textbf{33.41}& \textbf{20.28}& \textbf{13.96}& \textbf{10.11}& \textbf{30.56}& \textbf{42.23} \\ \bottomrule
    \end{tabular}
}
\caption{Performance comparison of translation results on the How2Sign dataset. YT-ASL-SLT and SSVP-SLT (marked with \(\dagger\)) are reported without dataset scaling to ensure a fair comparison.}
\label{tab:res_h2s}
\vspace{-1.5em}
\end{table*}

\section{Experiments}

\subsection{Implementation Details}

For SE and ME, we use CLIP ViT-L/14~\cite{radford2021learning} and VideoMAE-L/16~\cite{tong2022videomae}, respectively. To extract the spatial features, the sign images are interpolated to multiple scales, such as \(224 \times 224\) and \(448 \times 448\). For each scale, larger images are split into sub-images of regular size (\(224 \times 224\)) and processed individually. These features from the sub-images are then pooled and concatenated with features from the original representation. 
For the motion features, each clip consists of 16 frames, based on the findings from~\cite{wilbur2009effects}, which suggests that this frame interval captures a single sign. We set the stride \(s\) between consecutive clips to 8. 
We use FlanT5-XL~\cite{chung2024scaling} as the LLM for PHOENIX14T and How2Sign, while mT0-XL~\cite{muennighoff2022crosslingual} is used for the CSL-Daily.
During the warm-up phase with VT-Align, we use 4K steps on PHOENIX14T and CSL-Daily, and 15K steps on How2Sign.
Additional implementation details can be found in Appendix~\ref{sec:more_imple_detail}.

\subsection{Experimental Settings}\label{sec:exp_settings}

\paragraph{Datasets.} 
We evaluated our method on three sign language datasets: 
PHOENIX14T~\cite{camgoz2018neural}, CSL-Daily~\cite{zhou2021improving}, and How2Sign~\cite{duarte2021how2sign}. 
\textbf{PHOENIX14T} is a German Sign Language dataset comprising 8,257 samples and a vocabulary of 2,887 German words. 
\textbf{CSL-Daily} is a Chinese Sign Language dataset with 20,654 samples and a 2,343 Chinese characters.
\textbf{How2Sign} focuses on American Sign Language and includes 35,191 samples with a vocabulary of 16K English words. Detailed dataset statistics are provided in Appendix~\ref{sec:ds_stats}.

\paragraph{Evaluation Metrics.}
We report BLEU via SacreBLEU~\cite{papineni2002bleu,post2018call}\footnote{\scriptsize\texttt{nrefs:1|case:mixed|eff:no|tok:13a/zh|smooth:exp|version:2.2.1}} and ROUGE-L~\cite{lin2004automatic}. BLEU-n assesses translation precision by evaluating n-grams. ROUGE-L measures text similarity by calculating the F1 score based on the longest common subsequences between predicted and reference texts. We also report BLEURT~\cite{sellam2020bleurt} from the BLEURT-20 checkpoint\footnote{\scriptsize\texttt{\url{https://huggingface.co/lucadiliello/BLEURT-20}}}, which has been shown to correlate well with human judgments.

\paragraph{A Taxonomy of SLT.} 
In Section~\ref{sec:related_work}, we explored gloss-free methods, including those that incorporate gloss-supervised visual encoders. Although these approaches have traditionally been categorized as gloss-free, we argue that they should more accurately be described as \textit{weakly gloss-free} due to their dependence on gloss-annotated data. This classification is detailed in Table~\ref{tab:res_p14t_csl}. Specifically, methods such as TSPNet~\cite{li2020tspnet}, GASLT~\cite{yin2023gloss}, ConSLT~\cite{fu2023token}, GloFE-VN~\cite{lin2023gloss}, and OpenSLT~\cite{tarres2023sign} rely on sign features extracted by visual encoders trained on continuous or isolated sign language recognition (SLR) datasets.

\subsection{Comparison with State-of-the-Art}\label{sec:main_res}

\paragraph{Results on PHOENIX14T and CSL-Daily.} 
We first compared our method with both gloss-based and gloss-free methods on PHOENIX14T. As shown in Table~\ref{tab:res_p14t_csl}, most previous methods rely on the domain-specific fine-tuning of their visual encoders. By contrast, our method demonstrates consistent improvements across all reported metrics on PHOENIX14T without such fine-tuning. The only exception is ROUGE, where we achieved the second-best result. 
Specifically, the improvement on BLEU-4 is by a margin of 0.92, representing a 3.93\% increase over SignLLM~\cite{gong2024llms}. 
On CSL-Daily, which covers a broader range of topics than PHOENIX14T, the performance gains are even more pronounced. Our method achieved a margin increase of 4.8 in BLEU-4, reflecting a 30.41\% improvement over SignLLM.

\paragraph{Results on How2Sign.} 
Next, we evaluated our method on How2Sign, which poses greater challenges than PHOENIX14T due to its broader open-domain nature, longer sign videos, and larger vocabulary. As shown in Table~\ref{tab:res_h2s}, our method outperformed previous methods across all reported metrics. Specifically, we achieved a 0.45 margin in BLEU-4 which represents a 4.66\% improvement over Fla-LLM~\cite{chen2024factorized}. We see a performance gain in BLEURT, reaching 2.93, which is 7.46\% higher than SSVP-SLT~\cite{rust2024towards} under the non-scaled dataset setting. 

\begin{table}[t]
\scriptsize
\centering
\resizebox{\linewidth}{!}{%
    \renewcommand{\arraystretch}{0.95}
    \begin{tabular} {ccc  ccccc} \toprule
    \multicolumn{3}{c}{\textbf{Component}} & \multicolumn{5}{c}{\textbf{Metric}} \\ \cmidrule(lr){1-3} \cmidrule(lr){4-8}
    \textbf{SE}& \textbf{ME}& \textbf{VT-Align}& \textbf{B1}& \textbf{B2}& \textbf{B3}& \textbf{B4}& \textbf{RG} \\ \midrule
    \(\checkmark\)& & & 46.44& 33.79& 26.07& 21.11& 42.15 \\
    & \(\checkmark\)& & 29.71& 16.23& 10.99& 8.36& 22.44 \\
    \(\checkmark\)& \(\checkmark\)& & 47.59& 35.05& 27.34& 22.26& 43.92 \\
    \(\checkmark\)& & \(\checkmark\)& 48.12& 35.19& 27.42& 22.49& 44.19 \\
    \noalign{\vskip 0.3ex}\cdashline{1-8}\noalign{\vskip 0.7ex}
    \(\checkmark\)& \(\checkmark\)& \(\checkmark\)& \textbf{49.80}& \textbf{37.32} &\textbf{29.50} &\textbf{24.32} &\textbf{46.57} \\ \bottomrule
    \end{tabular}
}
\caption{Ablation study of the main component.}\label{tab:component_effect}
\end{table}
\begin{table}[t]
\scriptsize
\centering
\resizebox{\linewidth}{!}{%
    \renewcommand{\arraystretch}{0.95}
    \begin{tabular} {l cc c} \toprule
        \textbf{Models}& \textbf{\#Trainable Params}& \textbf{\#Total Params}& \textbf{B4} \\ \midrule
        w/o LLM& 60.5M& 60.5M& 6.35 \\
        mBART-L~\cite{liu2020multilingual}& 680M& 680M& 10.94 \\
        mT0-XL~\cite{muennighoff2022crosslingual}& 23.5M& 3.5B& 24.23 \\
        Llama-2~\cite{touvron2023llama}& 32.4M& 7B& 13.86 \\ 
        \noalign{\vskip 0.3ex}\cdashline{1-4}\noalign{\vskip 0.7ex}
        Flan-T5-XL~\cite{chung2024scaling}& 22.7M& 3B& \textbf{24.32} \\ \bottomrule
    \end{tabular}
}
\caption{Ablation study of the impact of LLM.}
\label{tab:abl_llm}
\vspace{-1.5em}
\end{table}

\begin{table}[t]
\scriptsize
\centering
\resizebox{.9\linewidth}{!}{%
    \renewcommand{\arraystretch}{0.95}
    \begin{tabular} {l c} \toprule
    \textbf{Method}& \textbf{KDEs Entropy} \(\downarrow\) \\ \midrule
    GFSLT-VLP~\cite{zhou2023gloss}& 0.32 \\
    \textbf{SPaMo (Ours)}& \textbf{0.12} \\ \bottomrule
    \end{tabular}
}
\caption{Comparison of KDE entropy values across different embeddings. Lower entropy values indicate more confident and distinct representations.}
\label{tab:abl_kde}
\vspace{-1.5em}
\end{table}
\begin{figure}[t]
    \centering
    \includegraphics[trim=0cm 0.5cm 0cm 0cm,clip=true,width=\linewidth]{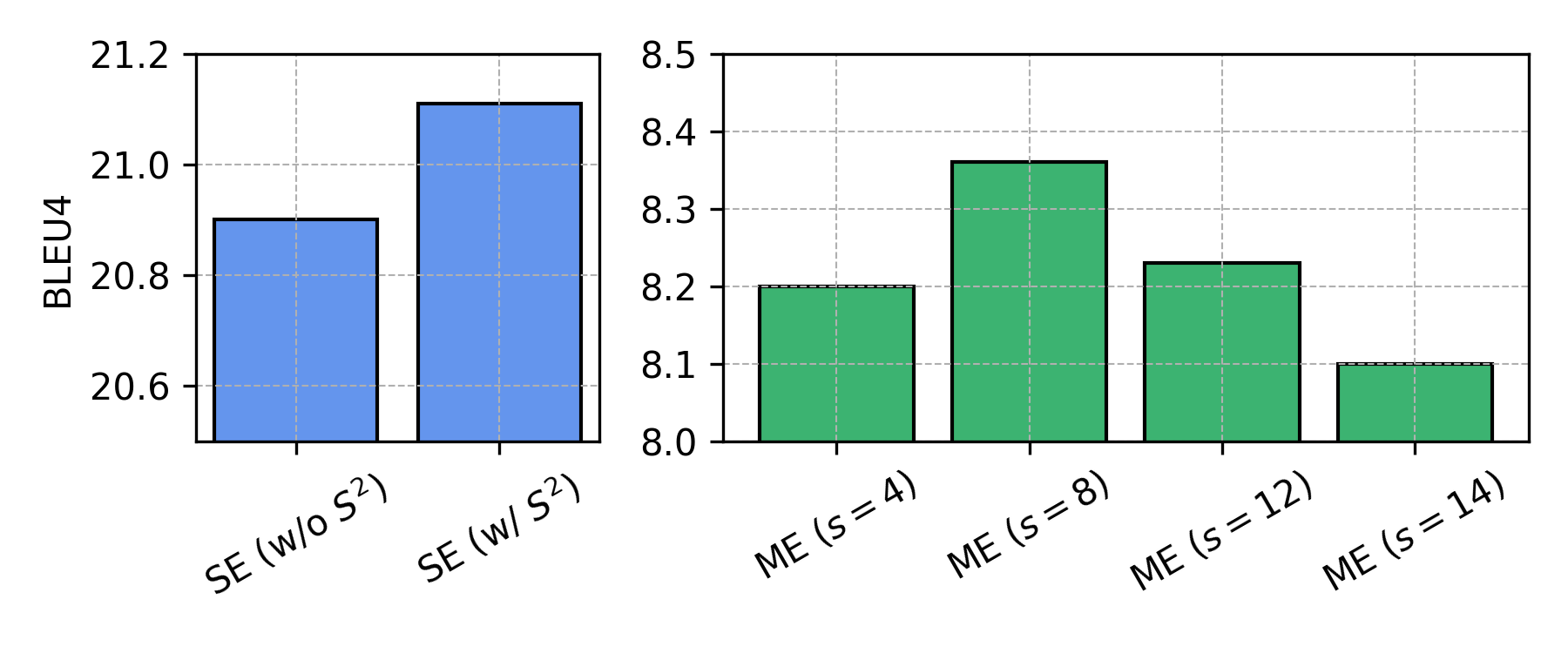}
    \vspace{-1.5em}
    \caption{Ablation study for SE and ME. \(S^2\) represents Scaling on Scales, and \(s\) denotes stride size. Note that the presented results do not include VT-Align.} 
    \label{fig:abl_s2_neigh_gap}
    \vspace{-1.5em}
\end{figure}

\paragraph{Kernel Density Estimation.}
To assess the quality of sign representations, following~\citet{ye2023cross}, we employed the Kernel Density Estimation (KDE) to estimate the probability density functions of embeddings from GFSLT-VLP and our method on PHOENIX14T. Note that we reproduced GFSLT-VLP using the official code\footnote{\scriptsize\texttt{\url{https://github.com/zhoubenjia/GFSLT-VLP}}}. As shown in Table~\ref{tab:abl_kde}, our method produced more compact and confident representations than GFSLT-VLP. More details on the KDE process are provided in Appendix~\ref{sec:appen_kde}.

\subsection{Ablation Study}

To further evaluate our method, we conducted extensive ablation experiments on PHOENIX14T, the most widely used sign language dataset. Additional results can be found in Appendix~\ref{sec:add_abl_res}.

\paragraph{Effect of Main Components.} 
We begin by evaluating the impact of the key components in our framework: SE, ME, and VT-Align. 
As shown in Table~\ref{tab:component_effect}, using SE or ME individually results in lower performance, with ME performing the worst in terms of BLEU-4. However, combining SE and ME leads to an overall improvement. Notably, when VT-Align is integrated with SE, the performance rises, nearly matching Sign2GPT (22.52 vs. 22.49). The best results are achieved when all components (SE, ME, and VT-Align) are used together, yielding the highest scores across all metrics. 
This highlights the importance of each component in enhancing overall performance of SpaMo.

\paragraph{Effect of LLM.}
Next, we explored the impact of different types of LLMs by replacing our model, as shown in Table~\ref{tab:abl_llm}. 
We compared five models, each with a different number of parameters: our method without pre-trained weights, mBART-L, mT0-XL, Flan-T5-XL, and Llama-2. Among these, Flan-T5-XL achieves the best performance, though it nearly ties with mT0-XL. Interestingly, despite its larger parameter count, Llama-2, which is also employed in SignLLM, does not outperform the others. 
This finding aligns with the observations of~\citeauthor{zhang2024scaling}, suggesting that scaling up LMs does not always lead to better performance. 
In our case, the main reason likely lies in the fact that larger models generally demand more extensive and higher-quality data to unlock their full potential~\cite{Kaplan2020ScalingLF,hoffmann2022training}. Since PHOENIX14T is constrained in both scale and diversity, merely increasing the model size does not necessarily yield substantial performance gains.

\paragraph{Effect of \(S^2\) and Neighboring Gap.} 
Finally, we evaluated the effect of \(S^2\) and the gap between neighboring clips on SE and ME, respectively. 
As shown in Figure~\ref{fig:abl_s2_neigh_gap}, \(S^2\) substantially improves translation performance, highlighting its effectiveness in helping SE capture more spatial details. 
Additionally, our analysis reveals that a stride size of 8 between neighboring clips yields the best results, suggesting that this stride size optimally aids ME in extracting the motion dynamics.

\begin{table}[t]
    \scriptsize
    \centering
    \resizebox{\linewidth}{!}{%
        \begin{tabular} {rl} \toprule
            \multirow{2}{*}{Ref:}& die neue woche beginnt noch wechselhaft und etwas kühler. \\
                                & \textit{(the new week begins still changeable and somewhat cooler)} \\
            \multirow{2}{*}{GFSLT-VLP:}& \textcolor{red}{am montag} wieder \textcolor{blue}{wechselhaft und} \textcolor{blue}{kühler}.  \\
                                & \textit{(on Monday again changeable and cooler)} \\
            \multirow{2}{*}{Ours:}& \textcolor{blue}{die} \textcolor{blue}{neue} \textcolor{blue}{woche} \textcolor{blue}{beginnt} \textcolor{blue}{wechselhaft und} wieder \textcolor{blue}{kühler}. \\
                                & \textit{(the new week begins changeable and again cooler)} \\ \midrule

            \multirow{2}{*}{Ref:}& sonst viel sonnenschein. \\
                    & \textit{otherwise, a lot of sunshine.} \\
            \multirow{2}{*}{GFSLT-VLP:}& \textcolor{red}{im übrigen land} \textcolor{blue}{viel} sonne. \\
                                & \textit{in the rest of the country, a lot of sun.} \\
            \multirow{2}{*}{Ours:}& \textcolor{blue}{sonst} \textcolor{blue}{viel} \textcolor{blue}{sonnenschein}. \\
                                & \textit{otherwise, a lot of sunshine.} \\ \bottomrule
        \end{tabular}
    }
    \vspace{-0.5em}
    \caption{Translation results on the test set compared to GFSLT-VLP on PHOENIX14T. Correctly translated 1-grams are highlighted in \textcolor{blue}{blue}, while incorrect translations are marked in \textcolor{red}{red}.}
    \label{tab:qaul_res_p14t_main}
    \vspace{-1em}
\end{table}
\begin{table}[t]
    \scriptsize
    \centering
    \resizebox{\linewidth}{!}{%
        \begin{tabular} {rl} \toprule

            \multirow{2}{*}{Vis. Token:}& \colorbox{green!40}{NORDWEST} SONST \colorbox{green!40}{FREUNDLICH} STURDY \\
                                        & \textit{(NORTHWEST OTHERWISE FRIENDLY STURDY)} \\
            \multirow{2}{*}{Gloss:}& \colorbox{green!40}{NORDWEST} \colorbox{green!40}{FREUNDLICH} \\
                                    & \textit{(NORTHWEST FRIENDLY)} \\
            \multirow{2}{*}{Translation:}&  richtung norden und westen ist es recht freundlich. \\ 
                                        & \textit{(Towards the north and west it is quite pleasant.)} \\ \midrule

            \multirow{2}{*}{Vis. Token:}& \colorbox{pink!40}{BLEIBT} \colorbox{green!40}{WIND} WINTER \\
                                        & \textit{(REMAINS WIND WINTER)} \\
            \multirow{2}{*}{Gloss:}& \colorbox{pink!40}{BLEIBEN} \colorbox{green!40}{WIND} \\ 
                                    & \textit{(REMAIN WIND)} \\
            \multirow{2}{*}{Translation:}&  es bleibt windig. \\ 
                                        & \textit{(it remains windy.)} \\ \midrule

            \multirow{2}{*}{Vis. Token:}& \colorbox{blue!40}{LIEBE} \colorbox{pink!40}{GUTEN} \colorbox{green!40}{ABEND} SCHÖNEN \\
                                        & \textit{(DEAR GOOD EVENING BEAUTIFUL)} \\
            \multirow{2}{*}{Gloss:}& \colorbox{pink!40}{GUT} \colorbox{green!40}{ABEND} BEGRUESSEN \\ 
                                    & \textit{(GOOD EVENING GREETINGS)} \\
            \multirow{2}{*}{Translation:}&  guten abend \colorbox{blue!40}{liebe} zuschauer. \\ 
                                        & \textit{(good evening dear viewers.)} \\ \bottomrule
        \end{tabular}
    }
    \vspace{-0.5em}
    \caption{\fboxsep1.5pt Comparison between visual tokens (Vis. Token) and their corresponding glosses. Words highlighted in \colorbox{green!40}{green} are exact matches, those in \colorbox{pink!40}{pink} are semantic matches, and words in \colorbox{blue!40}{blue} are absent in the gloss but appear in the translation.}
    \label{tab:vistok_res_p14t}
    \vspace{-2em}
\end{table}

\begin{figure}[t]
    \centering
    \includegraphics[trim=2cm 1.5cm 2cm 2cm,clip=true,width=\linewidth]{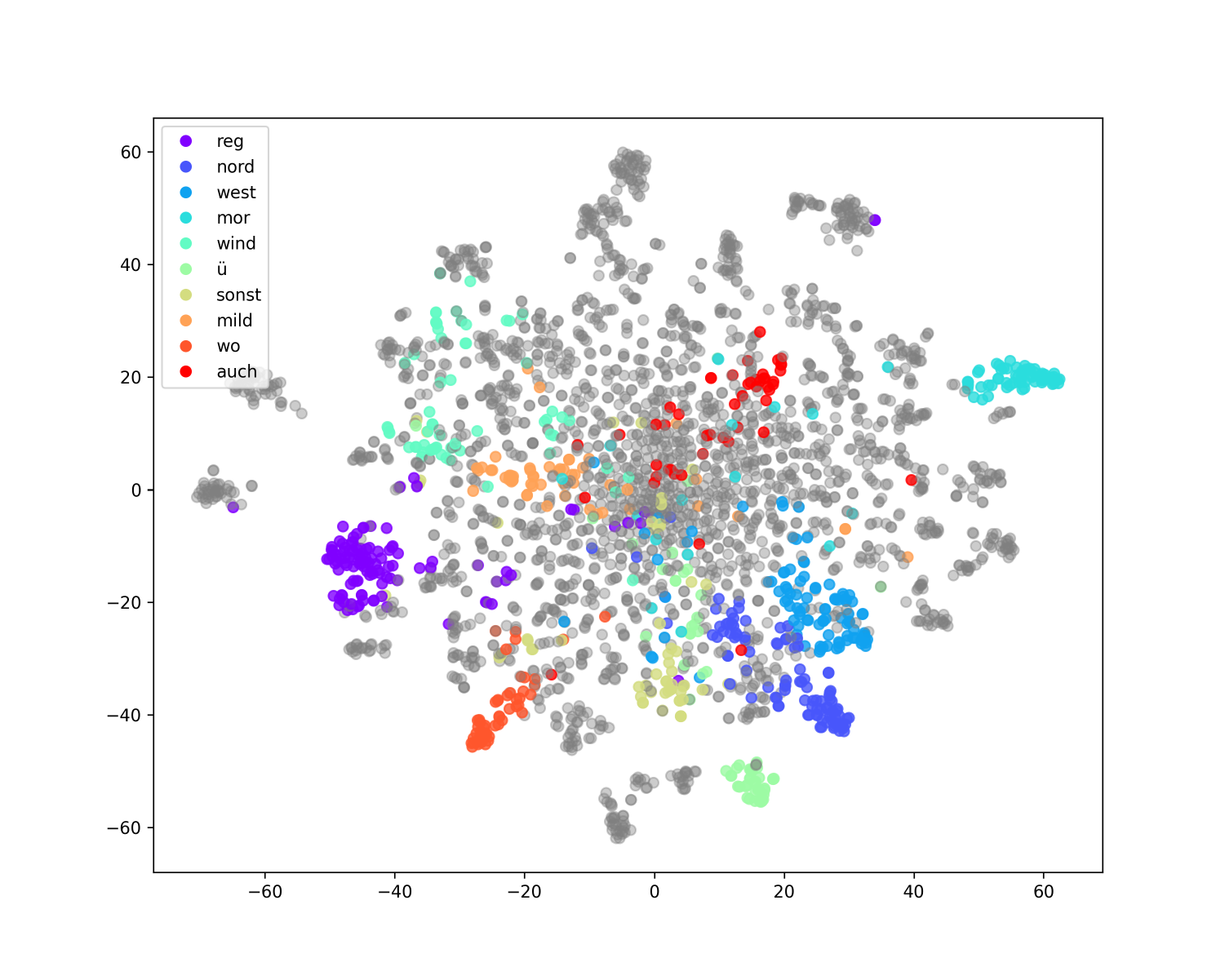}
    \vspace{-1.5em}
    \caption{The t-SNE visualization of sign features. Different colors represent features with distinct semantics, while gray points are other categories not listed.} 
    \label{fig:tsne_vis_emb_p14t}
    \vspace{-1.5em}
\end{figure}

\subsection{Qualitative Analysis}

\paragraph{Translation Results.}
Table~\ref{tab:qaul_res_p14t_main} presents two example translations on PHOENIX14T, comparing our method with GFSLT-VLP, the only other publicly available baseline. In the first example (top), our method provides an accurate translation, whereas GFSLT-VLP fails to capture the correct semantic meaning. In the second example (bottom), our method again produces a precise translation, while GFSLT-VLP introduces errors, resulting in incorrect information. These examples demonstrate the superior accuracy of our method in generating reliable translations. More translation examples are shown in Appendix~\ref{sec:add_qual_res}.

\paragraph{Visual Token Analysis.}
We performed an additional analysis to explore how the LLM interprets the sign videos. 
For each visual feature, we identified the word with the shortest distance in the LLM's embedding space, representing the closest match. Further details can be found in Appendix~\ref{sec:append_gen_vis_tok}.
Figure~\ref{fig:tsne_vis_emb_p14t} shows the t-SNE visualization of each sign feature mapped to the corresponding word. 
We observed that certain visual features align closely with specific words, which likely represents the semantic concepts that the LLM associates with these features. 
In other words, these words represent the LLM's interpretation or labeling of the visual content. 
We refer to these mapped words as ``visual tokens''. 
We further compared these visual tokens with the ground-truth glosses as shown in Table~\ref{tab:vistok_res_p14t}. To ensure a clearer and more accurate semantic comparison, repetitive words were removed from the visual tokens. 
Surprisingly, the LLM's interpretation of the sign videos is similar to the glosses, though not perfectly aligned. This suggests that the LLM has learned to link particular video patterns with specific textual concepts, explaining why those words cluster near the visual features in the embedding space. Additionally, we found that the visual tokens capture words that are present in the translation but not in the glosses. This finding suggests that visual tokens may provide a more comprehensive representation than current glosses, potentially broadening their scope beyond what has been traditionally documented.

\section{Conclusion}

In this paper, we introduced SpaMo, a novel gloss-free SLT framework based on LLMs. Apart from the previous methods that rely on domain-specific fine-tuning of their visual encoders, SpaMo focuses on capturing the spatial configurations and motion dynamics, eliminating the need for resource-intensive fine-tuning. 
We also proposed VT-Align, a training strategy that effectively aligns and narrows the modality gap between the sign videos and target texts, enabling the transformation of the sign videos into inputs interpretable by the LLM. 
Our approach achieved state-of-the-art results on three popular datasets. Furthermore, we provided the first comprehensive analysis of how the LLM interprets the sign videos within its embedding space and translates them into corresponding text.

\section*{Limitations}


Recently, scaling datasets~\cite{uthus2024youtube,rust2024towards} has consistently led to performance improvements, as seen with larger sign language datasets, such as Youtube-ASL~\cite{uthus2024youtube} and BOBSL~\cite{albanie2021bobsl}. While dataset scaling could also enhance our method, in this work, we focus on a constrained setting. 
Specifically, we use limited sign language datasets to evaluate and compare results, demonstrating the effectiveness of our method in resource-limited scenarios. Future work will involve expanding the dataset size to explore the full potential of our method and to assess its scalability and performance across more extensive and diverse datasets.

In this paper, we highlight that domain-specific fine-tuning of visual encoders is not essential for our method.
However, this implies that our method relies on visual encoders pre-trained on general tasks such as action recognition and image captioning. 
To bridge this gap, we introduce a pre-alignment process and apply LoRA fine-tuning to the LLM. 
While this might appear to be a compromise, it significantly reduces the resource requirements compared to fine-tuning both the visual encoders and the LLM. 
Additionally, as we discussed in the previous paragraph, this limitation can be addressed as more data becomes available, allowing for improved scalability and performance over time.

\section*{Ethics Statement}

Our work focuses on developing a practical framework for SLT with the goal of overcoming communication barriers faced by the Deaf and hard-of-hearing communities. Although our approach utilizes off-the-shelf visual encoders and LLMs, there is a possibility that the framework could produce unexpected or biased outputs due to the inherent limitations in the pre-trained models. However, we are optimistic that future advancements in LLMs will help mitigate these issues. We rely on open datasets such as PHOENIX14T~\cite{camgoz2018neural}, CSL-Daily~\cite{zhou2021improving}, and How2Sign~\cite{duarte2021how2sign}, which, while containing potentially identifiable information, present minimal concerns regarding personal privacy. 
Additionally, our method has been validated only on German, Chinese, and American sign languages, limiting its applicability to other sign languages. We call for future research in SLT to expand on a broader range of sign languages, promoting greater equity for the Deaf community.

\section*{Acknowledgment}
This work was supported by the Institute for Information and communications Technology Promotion (IITP) grant funded by the Korea government (MSIT) (No. 2022-0-00010, Development of Korean sign language translation service technology for the deaf in medical environment).


\bibliography{custom}

\begin{thebibliography}{72}
\providecommand{\natexlab}[1]{#1}

\bibitem[{Alayrac et~al.(2022)Alayrac, Donahue, Luc, Miech, Barr, Hasson, Lenc, Mensch, Millican, Reynolds et~al.}]{alayrac2022flamingo}
Jean-Baptiste Alayrac, Jeff Donahue, Pauline Luc, Antoine Miech, Iain Barr, Yana Hasson, Karel Lenc, Arthur Mensch, Katherine Millican, Malcolm Reynolds, et~al. 2022.
\newblock \href {https://proceedings.neurips.cc/paper_files/paper/2022/file/960a172bc7fbf0177ccccbb411a7d800-Paper-Conference.pdf} {Flamingo: a visual language model for few-shot learning}.
\newblock \emph{Advances in neural information processing systems}, 35:23716--23736.

\bibitem[{Albanie et~al.(2021)Albanie, Varol, Momeni, Bull, Afouras, Chowdhury, Fox, Woll, Cooper, McParland, and Zisserman}]{albanie2021bobsl}
Samuel Albanie, G{\"u}l Varol, Liliane Momeni, Hannah Bull, Triantafyllos Afouras, Himel Chowdhury, Neil Fox, Bencie Woll, Rob Cooper, Andrew McParland, and Andrew Zisserman. 2021.
\newblock \href {https://www.robots.ox.ac.uk/~vgg/data/bobsl} {{BOBSL}: {BBC}-{O}xford {B}ritish {S}ign {L}anguage {D}ataset}.
\newblock \emph{arXiv}.

\bibitem[{Bardes et~al.(2024)Bardes, Garrido, Ponce, Chen, Rabbat, LeCun, Assran, and Ballas}]{bardes2024revisiting}
Adrien Bardes, Quentin Garrido, Jean Ponce, Xinlei Chen, Michael Rabbat, Yann LeCun, Mahmoud Assran, and Nicolas Ballas. 2024.
\newblock \href {https://arxiv.org/pdf/2404.08471} {Revisiting feature prediction for learning visual representations from video}.
\newblock \emph{arXiv preprint arXiv:2404.08471}.

\bibitem[{Bosworth et~al.(2019)Bosworth, Wright, and Dobkins}]{bosworth2019analysis}
Rain~G Bosworth, Charles~E Wright, and Karen~R Dobkins. 2019.
\newblock \href {https://www.sciencedirect.com/science/article/am/pii/S0042698919301622} {Analysis of the visual spatiotemporal properties of american sign language}.
\newblock \emph{Vision research}, 164:34--43.

\bibitem[{Brown et~al.(2020)Brown, Mann, Ryder, Subbiah, Kaplan, Dhariwal, Neelakantan, Shyam, Sastry, Askell et~al.}]{brown2020language}
Tom Brown, Benjamin Mann, Nick Ryder, Melanie Subbiah, Jared~D Kaplan, Prafulla Dhariwal, Arvind Neelakantan, Pranav Shyam, Girish Sastry, Amanda Askell, et~al. 2020.
\newblock \href {https://proceedings.neurips.cc/paper_files/paper/2020/file/1457c0d6bfcb4967418bfb8ac142f64a-Paper.pdf} {Language models are few-shot learners}.
\newblock \emph{Advances in neural information processing systems}, 33:1877--1901.

\bibitem[{Camgoz et~al.(2018)Camgoz, Hadfield, Koller, Ney, and Bowden}]{camgoz2018neural}
Necati~Cihan Camgoz, Simon Hadfield, Oscar Koller, Hermann Ney, and Richard Bowden. 2018.
\newblock \href {http://openaccess.thecvf.com/content_cvpr_2018/papers/Camgoz_Neural_Sign_Language_CVPR_2018_paper.pdf} {Neural sign language translation}.
\newblock In \emph{Proceedings of the IEEE conference on computer vision and pattern recognition}, pages 7784--7793.

\bibitem[{Camgoz et~al.(2020)Camgoz, Koller, Hadfield, and Bowden}]{camgoz2020sign}
Necati~Cihan Camgoz, Oscar Koller, Simon Hadfield, and Richard Bowden. 2020.
\newblock \href {https://openaccess.thecvf.com/content_CVPR_2020/papers/Camgoz_Sign_Language_Transformers_Joint_End-to-End_Sign_Language_Recognition_and_Translation_CVPR_2020_paper.pdf} {Sign language transformers: Joint end-to-end sign language recognition and translation}.
\newblock In \emph{Proceedings of the IEEE/CVF conference on computer vision and pattern recognition}, pages 10023--10033.

\bibitem[{Chen et~al.(2022{\natexlab{a}})Chen, Wei, Sun, Wu, and Lin}]{chen2022simple}
Yutong Chen, Fangyun Wei, Xiao Sun, Zhirong Wu, and Stephen Lin. 2022{\natexlab{a}}.
\newblock \href {http://openaccess.thecvf.com/content/CVPR2022/papers/Chen_A_Simple_Multi-Modality_Transfer_Learning_Baseline_for_Sign_Language_Translation_CVPR_2022_paper.pdf} {A simple multi-modality transfer learning baseline for sign language translation}.
\newblock In \emph{Proceedings of the IEEE/CVF conference on computer vision and pattern recognition}, pages 5120--5130.

\bibitem[{Chen et~al.(2022{\natexlab{b}})Chen, Zuo, Wei, Wu, Liu, and Mak}]{chen2022two}
Yutong Chen, Ronglai Zuo, Fangyun Wei, Yu~Wu, Shujie Liu, and Brian Mak. 2022{\natexlab{b}}.
\newblock \href {https://proceedings.neurips.cc/paper_files/paper/2022/file/6cd3ac24cdb789beeaa9f7145670fcae-Paper-Conference.pdf} {Two-stream network for sign language recognition and translation}.
\newblock \emph{Advances in Neural Information Processing Systems}, 35:17043--17056.

\bibitem[{Chen et~al.(2024)Chen, Zhou, Li, Wan, Lei, Jiang, Lu, and Zhao}]{chen2024factorized}
Zhigang Chen, Benjia Zhou, Jun Li, Jun Wan, Zhen Lei, Ning Jiang, Quan Lu, and Guoqing Zhao. 2024.
\newblock \href {https://arxiv.org/pdf/2403.12556} {Factorized learning assisted with large language model for gloss-free sign language translation}.
\newblock In \emph{Proceedings of the 2024 Joint International Conference on Computational Linguistics, Language Resources and Evaluation (LREC-COLING 2024)}, pages 7071--7081. ELRA and ICCL.

\bibitem[{Cheng et~al.(2023)Cheng, Wei, Bao, Chen, and Zhang}]{cheng2023cico}
Yiting Cheng, Fangyun Wei, Jianmin Bao, Dong Chen, and Wenqiang Zhang. 2023.
\newblock \href {https://openaccess.thecvf.com/content/CVPR2023/papers/Bao_CiCo_Domain-Aware_Sign_Language_Retrieval_via_Cross-Lingual_Contrastive_Learning_CVPR_2023_paper.pdf} {Cico: Domain-aware sign language retrieval via cross-lingual contrastive learning}.
\newblock In \emph{Proceedings of the IEEE/CVF Conference on Computer Vision and Pattern Recognition}, pages 19016--19026.

\bibitem[{Cheng et~al.(2024)Cheng, Leng, Zhang, Xin, Li, Chen, Zhu, Zhang, Luo, Zhao et~al.}]{cheng2024videollama}
Zesen Cheng, Sicong Leng, Hang Zhang, Yifei Xin, Xin Li, Guanzheng Chen, Yongxin Zhu, Wenqi Zhang, Ziyang Luo, Deli Zhao, et~al. 2024.
\newblock \href {https://arxiv.org/pdf/2406.07476} {Videollama 2: Advancing spatial-temporal modeling and audio understanding in video-llms}.
\newblock \emph{arXiv preprint arXiv:2406.07476}.

\bibitem[{Chiang et~al.(2023)Chiang, Li, Lin, Sheng, Wu, Zhang, Zheng, Zhuang, Zhuang, Gonzalez et~al.}]{chiang2023vicuna}
Wei-Lin Chiang, Zhuohan Li, Zi~Lin, Ying Sheng, Zhanghao Wu, Hao Zhang, Lianmin Zheng, Siyuan Zhuang, Yonghao Zhuang, Joseph~E Gonzalez, et~al. 2023.
\newblock \href {https://vicuna. lmsys. org} {Vicuna: An open-source chatbot impressing gpt-4 with 90\%* chatgpt quality}.
\newblock 2(3):6.

\bibitem[{Chung et~al.(2024)Chung, Hou, Longpre, Zoph, Tay, Fedus, Li, Wang, Dehghani, Brahma et~al.}]{chung2024scaling}
Hyung~Won Chung, Le~Hou, Shayne Longpre, Barret Zoph, Yi~Tay, William Fedus, Yunxuan Li, Xuezhi Wang, Mostafa Dehghani, Siddhartha Brahma, et~al. 2024.
\newblock \href {https://www.jmlr.org/papers/volume25/23-0870/23-0870.pdf} {Scaling instruction-finetuned language models}.
\newblock \emph{Journal of Machine Learning Research}, 25(70):1--53.

\bibitem[{Duarte et~al.(2021)Duarte, Palaskar, Ventura, Ghadiyaram, DeHaan, Metze, Torres, and Giro-i Nieto}]{duarte2021how2sign}
Amanda Duarte, Shruti Palaskar, Lucas Ventura, Deepti Ghadiyaram, Kenneth DeHaan, Florian Metze, Jordi Torres, and Xavier Giro-i Nieto. 2021.
\newblock \href {https://openaccess.thecvf.com/content/CVPR2021/papers/Duarte_How2Sign_A_Large-Scale_Multimodal_Dataset_for_Continuous_American_Sign_Language_CVPR_2021_paper.pdf} {How2sign: a large-scale multimodal dataset for continuous american sign language}.
\newblock In \emph{Proceedings of the IEEE/CVF conference on computer vision and pattern recognition}, pages 2735--2744.

\bibitem[{Emmorey and Casey(1995)}]{emmorey1995comparison}
Karen Emmorey and Shannon Casey. 1995.
\newblock \href {https://www.semanticscholar.org/paper/A-Comparison-of-Spatial-Language-in-English-%26-Sign-Emmorey-Casey/4f40fdf6fe98d02fb3723f40c2398c6ce5b061f8} {A comparison of spatial language in english \& american sign language}.
\newblock \emph{Sign Language Studies}, 88(1):255--288.

\bibitem[{Feng et~al.(2024)Feng, Lin, Dwivedi, Sun, Patel, and Black}]{feng2024chatpose}
Yao Feng, Jing Lin, Sai~Kumar Dwivedi, Yu~Sun, Priyanka Patel, and Michael~J Black. 2024.
\newblock \href {https://openaccess.thecvf.com/content/CVPR2024/papers/Feng_ChatPose_Chatting_about_3D_Human_Pose_CVPR_2024_paper.pdf} {Chatpose: Chatting about 3d human pose}.
\newblock In \emph{Proceedings of the IEEE/CVF Conference on Computer Vision and Pattern Recognition}, pages 2093--2103.

\bibitem[{Fu et~al.(2023)Fu, Ye, Zhang, Yu, Hu, Shi, and Chen}]{fu2023token}
Biao Fu, Peigen Ye, Liang Zhang, Pei Yu, Cong Hu, Xiaodong Shi, and Yidong Chen. 2023.
\newblock \href {https://ieeexplore.ieee.org/iel7/10094559/10094560/10095466.pdf?casa_token=AgsE3cD_g3kAAAAA:oBozKzp-lKnGCeHvQ_zXYNuWFvXyXEtIrHOwVyLB1FZVCkn_-XSJGIC80oczAbmp6RPXzU8A} {A token-level contrastive framework for sign language translation}.
\newblock In \emph{ICASSP 2023-2023 IEEE International Conference on Acoustics, Speech and Signal Processing (ICASSP)}, pages 1--5. IEEE.

\bibitem[{Gao et~al.(2024)Gao, He, Wu, and Wang}]{gao2024towards}
Pengzhi Gao, Zhongjun He, Hua Wu, and Haifeng Wang. 2024.
\newblock \href {https://arxiv.org/pdf/2401.05861} {Towards boosting many-to-many multilingual machine translation with large language models}.
\newblock \emph{arXiv preprint arXiv:2401.05861}.

\bibitem[{Gong et~al.(2024)Gong, Foo, He, Rahmani, and Liu}]{gong2024llms}
Jia Gong, Lin~Geng Foo, Yixuan He, Hossein Rahmani, and Jun Liu. 2024.
\newblock \href {https://openaccess.thecvf.com/content/CVPR2024/papers/Gong_LLMs_are_Good_Sign_Language_Translators_CVPR_2024_paper.pdf} {Llms are good sign language translators}.
\newblock In \emph{Proceedings of the IEEE/CVF Conference on Computer Vision and Pattern Recognition}, pages 18362--18372.

\bibitem[{Hoffmann et~al.(2022)Hoffmann, Borgeaud, Mensch, Buchatskaya, Cai, Rutherford, Casas, Hendricks, Welbl, Clark et~al.}]{hoffmann2022training}
Jordan Hoffmann, Sebastian Borgeaud, Arthur Mensch, Elena Buchatskaya, Trevor Cai, Eliza Rutherford, Diego de~Las Casas, Lisa~Anne Hendricks, Johannes Welbl, Aidan Clark, et~al. 2022.
\newblock Training compute-optimal large language models.
\newblock \emph{arXiv preprint arXiv:2203.15556}.

\bibitem[{Hu et~al.(2021)Hu, Shen, Wallis, Allen-Zhu, Li, Wang, Wang, and Chen}]{hu2021lora}
Edward~J Hu, Yelong Shen, Phillip Wallis, Zeyuan Allen-Zhu, Yuanzhi Li, Shean Wang, Lu~Wang, and Weizhu Chen. 2021.
\newblock \href {https://arxiv.org/pdf/2106.09685)} {Lora: Low-rank adaptation of large language models}.
\newblock \emph{arXiv preprint arXiv:2106.09685}.

\bibitem[{Hu et~al.(2023)Hu, Gao, Liu, and Feng}]{hu2023continuous}
Lianyu Hu, Liqing Gao, Zekang Liu, and Wei Feng. 2023.
\newblock \href {https://openaccess.thecvf.com/content/CVPR2023/papers/Hu_Continuous_Sign_Language_Recognition_With_Correlation_Network_CVPR_2023_paper.pdf} {Continuous sign language recognition with correlation network}.
\newblock In \emph{Proceedings of the IEEE/CVF Conference on Computer Vision and Pattern Recognition}, pages 2529--2539.

\bibitem[{Huang et~al.(2024)Huang, Zhou, Yao, and Han}]{huang2024froster}
Xiaohu Huang, Hao Zhou, Kun Yao, and Kai Han. 2024.
\newblock \href {https://openreview.net/forum?id=zYXFMeHRtO} {{FROSTER:} frozen {CLIP} is {A} strong teacher for open-vocabulary action recognition}.
\newblock In \emph{The Twelfth International Conference on Learning Representations, {ICLR} 2024, Vienna, Austria, May 7-11, 2024}. OpenReview.net.

\bibitem[{Hwang et~al.(2024)Hwang, Cho, Lee, Yoon, and Park}]{hwang2024universal}
Eui~Jun Hwang, Sukmin Cho, Huije Lee, Youngwoo Yoon, and Jong~C Park. 2024.
\newblock \href {https://arxiv.org/pdf/2407.02854} {Universal gloss-level representation for gloss-free sign language translation and production}.
\newblock \emph{arXiv preprint arXiv:2407.02854}.

\bibitem[{Jia et~al.(2021)Jia, Yang, Xia, Chen, Parekh, Pham, Le, Sung, Li, and Duerig}]{jia2021scaling}
Chao Jia, Yinfei Yang, Ye~Xia, Yi-Ting Chen, Zarana Parekh, Hieu Pham, Quoc Le, Yun-Hsuan Sung, Zhen Li, and Tom Duerig. 2021.
\newblock \href {http://proceedings.mlr.press/v139/jia21b/jia21b.pdf} {Scaling up visual and vision-language representation learning with noisy text supervision}.
\newblock In \emph{International conference on machine learning}, pages 4904--4916. PMLR.

\bibitem[{Jing et~al.(2024)Jing, Song, Zu, Zheng, Zhao, and Nie}]{jing2024vk}
Liqiang Jing, Xuemeng Song, Xinxing Zu, Na~Zheng, Zhongzhou Zhao, and Liqiang Nie. 2024.
\newblock \href {https://ieeexplore.ieee.org/stamp/stamp.jsp?arnumber=10445887&casa_token=t5IcWST49jYAAAAA:pluBJ7uP7GdfwhNImLKVUbrLGrwx68dByW3PlZxRjrxdbbMEPhdehVhOLJ5c0Z15TrDopbp0mw&tag=1} {Vk-g2t: Vision and context knowledge enhanced gloss2text}.
\newblock In \emph{ICASSP 2024-2024 IEEE International Conference on Acoustics, Speech and Signal Processing (ICASSP)}, pages 7860--7864. IEEE.

\bibitem[{Ju et~al.(2023)Ju, Zheng, Wang, Zhao, and Liu}]{ju2023continuous}
Tianjie Ju, Yubin Zheng, Hanyi Wang, Haodong Zhao, and Gongshen Liu. 2023.
\newblock \href {https://aclanthology.org/2023.findings-acl.494.pdf} {Is continuous prompt a combination of discrete prompts? towards a novel view for interpreting continuous prompts}.
\newblock In \emph{Findings of the Association for Computational Linguistics: ACL 2023}, pages 7804--7819.

\bibitem[{Kaplan et~al.(2020)Kaplan, McCandlish, Henighan, Brown, Chess, Child, Gray, Radford, Wu, and Amodei}]{Kaplan2020ScalingLF}
Jared Kaplan, Sam McCandlish, Tom Henighan, Tom~B. Brown, Benjamin Chess, Rewon Child, Scott Gray, Alec Radford, Jeff Wu, and Dario Amodei. 2020.
\newblock \href {https://api.semanticscholar.org/CorpusID:210861095} {Scaling laws for neural language models}.
\newblock \emph{ArXiv}, abs/2001.08361.

\bibitem[{Li et~al.(2020{\natexlab{a}})Li, Rodriguez, Yu, and Li}]{li2020word}
Dongxu Li, Cristian Rodriguez, Xin Yu, and Hongdong Li. 2020{\natexlab{a}}.
\newblock \href {https://openaccess.thecvf.com/content_WACV_2020/papers/Li_Word-level_Deep_Sign_Language_Recognition_from_Video_A_New_Large-scale_WACV_2020_paper.pdf} {Word-level deep sign language recognition from video: A new large-scale dataset and methods comparison}.
\newblock In \emph{Proceedings of the IEEE/CVF winter conference on applications of computer vision}, pages 1459--1469.

\bibitem[{Li et~al.(2020{\natexlab{b}})Li, Xu, Yu, Zhang, Swift, Suominen, and Li}]{li2020tspnet}
Dongxu Li, Chenchen Xu, Xin Yu, Kaihao Zhang, Benjamin Swift, Hanna Suominen, and Hongdong Li. 2020{\natexlab{b}}.
\newblock \href {https://proceedings.neurips.cc/paper_files/paper/2020/file/8c00dee24c9878fea090ed070b44f1ab-Paper.pdf} {Tspnet: Hierarchical feature learning via temporal semantic pyramid for sign language translation}.
\newblock \emph{Advances in Neural Information Processing Systems}, 33:12034--12045.

\bibitem[{Li et~al.(2023)Li, Li, Savarese, and Hoi}]{li2023blip}
Junnan Li, Dongxu Li, Silvio Savarese, and Steven Hoi. 2023.
\newblock \href {https://proceedings.mlr.press/v202/li23q/li23q.pdf} {Blip-2: Bootstrapping language-image pre-training with frozen image encoders and large language models}.
\newblock In \emph{International conference on machine learning}, pages 19730--19742. PMLR.

\bibitem[{Lin and Och(2004)}]{lin2004automatic}
Chin-Yew Lin and Franz~Josef Och. 2004.
\newblock \href {https://aclanthology.org/P04-1077.pdf} {Automatic evaluation of machine translation quality using longest common subsequence and skip-bigram statistics}.
\newblock In \emph{Proceedings of the 42nd annual meeting of the association for computational linguistics (ACL-04)}, pages 605--612.

\bibitem[{Lin et~al.(2023)Lin, Wang, Zhu, Sun, Zhang, and Yang}]{lin2023gloss}
Kezhou Lin, Xiaohan Wang, Linchao Zhu, Ke~Sun, Bang Zhang, and Yi~Yang. 2023.
\newblock \href {https://arxiv.org/pdf/2305.12876} {Gloss-free end-to-end sign language translation}.
\newblock In \emph{Proceedings of the 61st Annual Meeting of the Association for Computational Linguistics}.

\bibitem[{Liu et~al.(2024{\natexlab{a}})Liu, Li, Li, and Lee}]{liu2024improved}
Haotian Liu, Chunyuan Li, Yuheng Li, and Yong~Jae Lee. 2024{\natexlab{a}}.
\newblock \href {https://openaccess.thecvf.com/content/CVPR2024/papers/Liu_Improved_Baselines_with_Visual_Instruction_Tuning_CVPR_2024_paper.pdf} {Improved baselines with visual instruction tuning}.
\newblock In \emph{Proceedings of the IEEE/CVF Conference on Computer Vision and Pattern Recognition}, pages 26296--26306.

\bibitem[{Liu et~al.(2024{\natexlab{b}})Liu, Li, Wu, and Lee}]{liu2024visual}
Haotian Liu, Chunyuan Li, Qingyang Wu, and Yong~Jae Lee. 2024{\natexlab{b}}.
\newblock \href {https://proceedings.neurips.cc/paper_files/paper/2023/file/6dcf277ea32ce3288914faf369fe6de0-Paper-Conference.pdf} {Visual instruction tuning}.
\newblock \emph{Advances in neural information processing systems}, 36.

\bibitem[{Liu et~al.(2020)Liu, Gu, Goyal, Li, Edunov, Ghazvininejad, Lewis, and Zettlemoyer}]{liu2020multilingual}
Yinhan Liu, Jiatao Gu, Naman Goyal, Xian Li, Sergey Edunov, Marjan Ghazvininejad, Mike Lewis, and Luke Zettlemoyer. 2020.
\newblock \href {https://doi.org/10.1162/TACL\_A\_00343} {Multilingual denoising pre-training for neural machine translation}.
\newblock \emph{Trans. Assoc. Comput. Linguistics}, 8:726--742.

\bibitem[{Loshchilov and Hutter(2017)}]{loshchilov2017decoupled}
Ilya Loshchilov and Frank Hutter. 2017.
\newblock \href {https://arxiv.org/pdf/1711.05101} {Decoupled weight decay regularization}.
\newblock \emph{arXiv preprint arXiv:1711.05101}.

\bibitem[{Muennighoff et~al.(2022)Muennighoff, Wang, Sutawika, Roberts, Biderman, Scao, Bari, Shen, Yong, Schoelkopf et~al.}]{muennighoff2022crosslingual}
Niklas Muennighoff, Thomas Wang, Lintang Sutawika, Adam Roberts, Stella Biderman, Teven~Le Scao, M~Saiful Bari, Sheng Shen, Zheng-Xin Yong, Hailey Schoelkopf, et~al. 2022.
\newblock \href {https://aclanthology.org/2023.acl-long.891.pdf} {Crosslingual generalization through multitask finetuning}.
\newblock In \emph{Annual Meeting of the Association for Computational Linguistics}, pages 15991--16111.

\bibitem[{Oquab et~al.(2023)Oquab, Darcet, Moutakanni, Vo, Szafraniec, Khalidov, Fernandez, Haziza, Massa, El-Nouby et~al.}]{oquab2023dinov2}
Maxime Oquab, Timoth{\'e}e Darcet, Th{\'e}o Moutakanni, Huy Vo, Marc Szafraniec, Vasil Khalidov, Pierre Fernandez, Daniel Haziza, Francisco Massa, Alaaeldin El-Nouby, et~al. 2023.
\newblock \href {https://arxiv.org/pdf/2304.07193} {Dinov2: Learning robust visual features without supervision}.
\newblock \emph{arXiv preprint arXiv:2304.07193}.

\bibitem[{Papineni et~al.(2002)Papineni, Roukos, Ward, and Zhu}]{papineni2002bleu}
Kishore Papineni, Salim Roukos, Todd Ward, and Wei-Jing Zhu. 2002.
\newblock \href {https://aclanthology.org/P02-1040.pdf} {Bleu: a method for automatic evaluation of machine translation}.
\newblock In \emph{Proceedings of the 40th annual meeting of the Association for Computational Linguistics}, pages 311--318.

\bibitem[{Post(2018)}]{post2018call}
Matt Post. 2018.
\newblock \href {https://aclanthology.org/W18-6319.pdf} {A call for clarity in reporting bleu scores}.
\newblock In \emph{Proceedings of the Third Conference on Machine Translation: Research Papers}.

\bibitem[{Radford et~al.(2021)Radford, Kim, Hallacy, Ramesh, Goh, Agarwal, Sastry, Askell, Mishkin, Clark et~al.}]{radford2021learning}
Alec Radford, Jong~Wook Kim, Chris Hallacy, Aditya Ramesh, Gabriel Goh, Sandhini Agarwal, Girish Sastry, Amanda Askell, Pamela Mishkin, Jack Clark, et~al. 2021.
\newblock \href {http://proceedings.mlr.press/v139/radford21a/radford21a.pdf} {Learning transferable visual models from natural language supervision}.
\newblock In \emph{International conference on machine learning}, pages 8748--8763. PMLR.

\bibitem[{Reimers and Gurevych(2019)}]{reimers2019sentence}
Nils Reimers and Iryna Gurevych. 2019.
\newblock \href {https://doi.org/10.18653/v1/D19-1410} {Sentence-{BERT}: Sentence embeddings using {S}iamese {BERT}-networks}.
\newblock In \emph{Proceedings of the 2019 Conference on Empirical Methods in Natural Language Processing and the 9th International Joint Conference on Natural Language Processing (EMNLP-IJCNLP)}, pages 3982--3992, Hong Kong, China. Association for Computational Linguistics.

\bibitem[{Rust et~al.(2024)Rust, Shi, Wang, Camg{\"o}z, and Maillard}]{rust2024towards}
Phillip Rust, Bowen Shi, Skyler Wang, Necati~Cihan Camg{\"o}z, and Jean Maillard. 2024.
\newblock \href {https://arxiv.org/pdf/2402.09611} {Towards privacy-aware sign language translation at scale}.
\newblock In \emph{Proceedings of the 62nd Annual Meeting of the Association for Computational Linguistics (Volume 1: Long Papers)}.

\bibitem[{Sellam et~al.(2020)Sellam, Das, and Parikh}]{sellam2020bleurt}
Thibault Sellam, Dipanjan Das, and Ankur~P Parikh. 2020.
\newblock \href {https://arxiv.org/pdf/2004.04696} {Bleurt: Learning robust metrics for text generation}.
\newblock In \emph{Proceedings of the 58th Annual Meeting of the Association for Computational Linguistics}, pages 7881--7892.

\bibitem[{Shi et~al.(2024)Shi, Wu, Mao, Wang, and Darrell}]{shi2024we}
Baifeng Shi, Ziyang Wu, Maolin Mao, Xin Wang, and Trevor Darrell. 2024.
\newblock \href {https://arxiv.org/pdf/2403.13043} {When do we not need larger vision models?}
\newblock \emph{arXiv preprint arXiv:2403.13043}.

\bibitem[{Shi et~al.(2022)Shi, Brentari, Shakhnarovich, and Livescu}]{shi2022open}
Bowen Shi, Diane Brentari, Greg Shakhnarovich, and Karen Livescu. 2022.
\newblock \href {https://arxiv.org/pdf/2205.12870} {Open-domain sign language translation learned from online video}.
\newblock In \emph{Proceedings of the Conference on Empirical Methods in Natural Language Processing}.

\bibitem[{Tang et~al.(2024)Tang, Zhao, Wen, and Liu}]{tang2024survey}
Zixuan Tang, Youjun Zhao, Yuhang Wen, and Mengyuan Liu. 2024.
\newblock \href {https://arxiv.org/pdf/2405.05584} {A survey on backbones for deep video action recognition}.
\newblock \emph{arXiv preprint arXiv:2405.05584}.

\bibitem[{Tarr{\'e}s et~al.(2023)Tarr{\'e}s, G{\'a}llego, Duarte, Torres, and Gir{\'o}-i Nieto}]{tarres2023sign}
Laia Tarr{\'e}s, Gerard~I G{\'a}llego, Amanda Duarte, Jordi Torres, and Xavier Gir{\'o}-i Nieto. 2023.
\newblock \href {https://openaccess.thecvf.com/content/CVPR2023W/WiCV/papers/Tarres_Sign_Language_Translation_from_Instructional_Videos_CVPRW_2023_paper.pdf} {Sign language translation from instructional videos}.
\newblock In \emph{Proceedings of the IEEE/CVF Conference on Computer Vision and Pattern Recognition (CVPR) Workshops}, pages 5625--5635.

\bibitem[{Tong et~al.(2022)Tong, Song, Wang, and Wang}]{tong2022videomae}
Zhan Tong, Yibing Song, Jue Wang, and Limin Wang. 2022.
\newblock \href {https://proceedings.neurips.cc/paper_files/paper/2022/file/416f9cb3276121c42eebb86352a4354a-Paper-Conference.pdf} {Videomae: Masked autoencoders are data-efficient learners for self-supervised video pre-training}.
\newblock \emph{Advances in neural information processing systems}, 35:10078--10093.

\bibitem[{Touvron et~al.(2023)Touvron, Martin, Stone, Albert, Almahairi, Babaei, Bashlykov, Batra, Bhargava, Bhosale et~al.}]{touvron2023llama}
Hugo Touvron, Louis Martin, Kevin Stone, Peter Albert, Amjad Almahairi, Yasmine Babaei, Nikolay Bashlykov, Soumya Batra, Prajjwal Bhargava, Shruti Bhosale, et~al. 2023.
\newblock \href {https://arxiv.org/pdf/2307.09288} {Llama 2: Open foundation and fine-tuned chat models}.
\newblock \emph{arXiv preprint arXiv:2307.09288}.

\bibitem[{Uthus et~al.(2024)Uthus, Tanzer, and Georg}]{uthus2024youtube}
Dave Uthus, Garrett Tanzer, and Manfred Georg. 2024.
\newblock \href {https://proceedings.neurips.cc/paper_files/paper/2023/file/5c61452daca5f0c260e683b317d13a3f-Paper-Datasets_and_Benchmarks.pdf} {Youtube-asl: A large-scale, open-domain american sign language-english parallel corpus}.
\newblock \emph{Advances in Neural Information Processing Systems}, 36.

\bibitem[{Wei and Chen(2023)}]{wei2023improving}
Fangyun Wei and Yutong Chen. 2023.
\newblock \href {https://openaccess.thecvf.com/content/ICCV2023/papers/Wei_Improving_Continuous_Sign_Language_Recognition_with_Cross-Lingual_Signs_ICCV_2023_paper.pdf} {Improving continuous sign language recognition with cross-lingual signs}.
\newblock In \emph{Proceedings of the IEEE/CVF International Conference on Computer Vision}, pages 23612--23621.

\bibitem[{Wilbur(2009)}]{wilbur2009effects}
Ronnie~B Wilbur. 2009.
\newblock \href {https://journals.sagepub.com/doi/pdf/10.1177/0023830909103174?casa_token=zVLoazTfb9wAAAAA:zgpEwi3raUpoacpi4yQzz-pVcftLvqtdQh7nh-XoIZjKIIOcKkQR6fSLqednkCt-dXQnsVIQFB0-} {Effects of varying rate of signing on asl manual signs and nonmanual markers}.
\newblock \emph{Language and speech}, 52(2-3):245--285.

\bibitem[{Wong et~al.(2024)Wong, Camgoz, and Bowden}]{wong2024sign2gpt}
Ryan Wong, Necati~Cihan Camgoz, and Richard Bowden. 2024.
\newblock \href {https://arxiv.org/pdf/2405.04164} {Sign2gpt: Leveraging large language models for gloss-free sign language translation}.
\newblock In \emph{Proceeding of the Eleventh International Conference on Learning Representations}.

\bibitem[{Yang et~al.(2023)Yang, Nagrani, Seo, Miech, Pont-Tuset, Laptev, Sivic, and Schmid}]{yang2023vid2seq}
Antoine Yang, Arsha Nagrani, Paul~Hongsuck Seo, Antoine Miech, Jordi Pont-Tuset, Ivan Laptev, Josef Sivic, and Cordelia Schmid. 2023.
\newblock \href {https://openaccess.thecvf.com/content/CVPR2023/papers/Yang_Vid2Seq_Large-Scale_Pretraining_of_a_Visual_Language_Model_for_Dense_CVPR_2023_paper.pdf} {Vid2seq: Large-scale pretraining of a visual language model for dense video captioning}.
\newblock In \emph{Proceedings of the IEEE/CVF Conference on Computer Vision and Pattern Recognition}, pages 10714--10726.

\bibitem[{Ye et~al.(2023)Ye, Jiao, Wang, Tu, and Xiong}]{ye2023cross}
Jinhui Ye, Wenxiang Jiao, Xing Wang, Zhaopeng Tu, and Hui Xiong. 2023.
\newblock \href {https://aclanthology.org/2023.findings-emnlp.904.pdf} {Cross-modality data augmentation for end-to-end sign language translation}.
\newblock In \emph{Findings of the Association for Computational Linguistics: EMNLP 2023}, pages 13558--13571.

\bibitem[{Yin et~al.(2021)Yin, Zhao, Liu, Jin, Zhang, Zeng, and He}]{yin2021simulslt}
Aoxiong Yin, Zhou Zhao, Jinglin Liu, Weike Jin, Meng Zhang, Xingshan Zeng, and Xiaofei He. 2021.
\newblock \href {https://dl.acm.org/doi/pdf/10.1145/3474085.3475544?casa_token=HPNVlsvXMggAAAAA:3-1RiKdO6KrcIsq6Kn3a_b1SqyAGxs7YIh0GSA9rMuLGTpaTXQ6mGmtIFYBjb5khojP7iXA-_Ahj} {Simulslt: End-to-end simultaneous sign language translation}.
\newblock In \emph{Proceedings of the 29th ACM International Conference on Multimedia}, pages 4118--4127.

\bibitem[{Yin et~al.(2023)Yin, Zhong, Tang, Jin, Jin, and Zhao}]{yin2023gloss}
Aoxiong Yin, Tianyun Zhong, Li~Tang, Weike Jin, Tao Jin, and Zhou Zhao. 2023.
\newblock \href {http://openaccess.thecvf.com/content/CVPR2023/papers/Yin_Gloss_Attention_for_Gloss-Free_Sign_Language_Translation_CVPR_2023_paper.pdf} {Gloss attention for gloss-free sign language translation}.
\newblock In \emph{Proceedings of the IEEE/CVF conference on computer vision and pattern recognition}, pages 2551--2562.

\bibitem[{Zhang et~al.(2023{\natexlab{a}})Zhang, Haddow, and Birch}]{zhang2023prompting}
Biao Zhang, Barry Haddow, and Alexandra Birch. 2023{\natexlab{a}}.
\newblock \href {https://proceedings.mlr.press/v202/zhang23m/zhang23m.pdf} {Prompting large language model for machine translation: A case study}.
\newblock In \emph{International Conference on Machine Learning}, pages 41092--41110. PMLR.

\bibitem[{Zhang et~al.(2023{\natexlab{b}})Zhang, M{\"u}ller, and Sennrich}]{zhang2023sltunet}
Biao Zhang, Mathias M{\"u}ller, and Rico Sennrich. 2023{\natexlab{b}}.
\newblock \href {https://openreview.net/forum?id=EBS4C77p_5S} {{SLTUNET}: A simple unified model for sign language translation}.
\newblock In \emph{The Eleventh International Conference on Learning Representations}.

\bibitem[{Zhang et~al.(2024{\natexlab{a}})Zhang, Tanzer, and Firat}]{zhang2024scaling}
Biao Zhang, Garrett Tanzer, and Orhan Firat. 2024{\natexlab{a}}.
\newblock \href {https://arxiv.org/pdf/2407.11855} {Scaling sign language translation}.
\newblock \emph{Advances in neural information processing systems}.

\bibitem[{Zhang et~al.(2024{\natexlab{b}})Zhang, Huang, Liu, Tang, Lu, Chen, Bai, Chu, Yu, and Ouyang}]{zhang2024motiongpt}
Yaqi Zhang, Di~Huang, Bin Liu, Shixiang Tang, Yan Lu, Lu~Chen, Lei Bai, Qi~Chu, Nenghai Yu, and Wanli Ouyang. 2024{\natexlab{b}}.
\newblock \href {https://ojs.aaai.org/index.php/AAAI/article/download/28567/29102} {Motiongpt: Finetuned llms are general-purpose motion generators}.
\newblock In \emph{Proceedings of the AAAI Conference on Artificial Intelligence}, 7, pages 7368--7376.

\bibitem[{Zhang et~al.(2024{\natexlab{c}})Zhang, Unell, Wang, Ghosh, Su, Schmidt, and Yeung-Levy}]{zhang2024visually}
Yuhui Zhang, Alyssa Unell, Xiaohan Wang, Dhruba Ghosh, Yuchang Su, Ludwig Schmidt, and Serena Yeung-Levy. 2024{\natexlab{c}}.
\newblock \href {https://arxiv.org/pdf/2405.18415} {Why are visually-grounded language models bad at image classification?}
\newblock \emph{arXiv preprint arXiv:2405.18415}.

\bibitem[{Zhao et~al.(2021)Zhao, Qi, Zhou, Duan, Zhou, and Li}]{zhao2021conditional}
Jian Zhao, Weizhen Qi, Wengang Zhou, Nan Duan, Ming Zhou, and Houqiang Li. 2021.
\newblock \href {https://ieeexplore.ieee.org/iel7/6046/4456689/09447976.pdf?casa_token=UdmNGZErgk0AAAAA:FBMCORXOICtBelgZheVquaPNPIPIgkcLB-D_AeywXa3T7c5Cuh_oancYcIo0otTOhMGRaFas} {Conditional sentence generation and cross-modal reranking for sign language translation}.
\newblock \emph{IEEE Transactions on Multimedia}, 24:2662--2672.

\bibitem[{Zhao et~al.(2024)Zhao, Zhang, Fu, Hu, Su, and Chen}]{zhao2024conditional}
Rui Zhao, Liang Zhang, Biao Fu, Cong Hu, Jinsong Su, and Yidong Chen. 2024.
\newblock \href {https://arxiv.org/abs/2312.15645} {Conditional variational autoencoder for sign language translation with cross-modal alignment}.
\newblock In \emph{Proceedings of the AAAI Conference on Artificial Intelligence}, volume~38, pages 19643--19651.

\bibitem[{Zhou et~al.(2023)Zhou, Chen, Clap{\'e}s, Wan, Liang, Escalera, Lei, and Zhang}]{zhou2023gloss}
Benjia Zhou, Zhigang Chen, Albert Clap{\'e}s, Jun Wan, Yanyan Liang, Sergio Escalera, Zhen Lei, and Du~Zhang. 2023.
\newblock \href {http://openaccess.thecvf.com/content/ICCV2023/papers/Zhou_Gloss-Free_Sign_Language_Translation_Improving_from_Visual-Language_Pretraining_ICCV_2023_paper.pdf} {Gloss-free sign language translation: Improving from visual-language pretraining}.
\newblock In \emph{Proceedings of the IEEE/CVF International Conference on Computer Vision}, pages 20871--20881.

\bibitem[{Zhou et~al.(2021{\natexlab{a}})Zhou, Zhou, Qi, Pu, and Li}]{zhou2021improving}
Hao Zhou, Wengang Zhou, Weizhen Qi, Junfu Pu, and Houqiang Li. 2021{\natexlab{a}}.
\newblock \href {http://openaccess.thecvf.com/content/CVPR2021/papers/Zhou_Improving_Sign_Language_Translation_With_Monolingual_Data_by_Sign_Back-Translation_CVPR_2021_paper.pdf} {Improving sign language translation with monolingual data by sign back-translation}.
\newblock In \emph{Proceedings of the IEEE/CVF Conference on Computer Vision and Pattern Recognition}, pages 1316--1325.

\bibitem[{Zhou et~al.(2021{\natexlab{b}})Zhou, Zhou, Zhou, and Li}]{zhou2021spatial}
Hao Zhou, Wengang Zhou, Yun Zhou, and Houqiang Li. 2021{\natexlab{b}}.
\newblock \href {https://ieeexplore.ieee.org/iel7/6046/4456689/09354538.pdf?casa_token=vqIW0tnoD7AAAAAA:lw5H_1S62WpSTRf9e6xFpX9Aaf3qxVUsI3hQipmxTKmydJKA_Ai_EwB1k3261H-ADmeEvLQ1} {Spatial-temporal multi-cue network for sign language recognition and translation}.
\newblock \emph{IEEE Transactions on Multimedia}, 24:768--779.

\bibitem[{Zhou et~al.(2024)Zhou, Arnab, Buch, Yan, Myers, Xiong, Nagrani, and Schmid}]{zhou2024streaming}
Xingyi Zhou, Anurag Arnab, Shyamal Buch, Shen Yan, Austin Myers, Xuehan Xiong, Arsha Nagrani, and Cordelia Schmid. 2024.
\newblock \href {https://openaccess.thecvf.com/content/CVPR2024/papers/Zhou_Streaming_Dense_Video_Captioning_CVPR_2024_paper.pdf} {Streaming dense video captioning}.
\newblock In \emph{Proceedings of the IEEE/CVF Conference on Computer Vision and Pattern Recognition}, pages 18243--18252.

\bibitem[{Zhu et~al.(2023)Zhu, Liu, Dong, Xu, Huang, Kong, Chen, and Li}]{zhu2023multilingual}
Wenhao Zhu, Hongyi Liu, Qingxiu Dong, Jingjing Xu, Shujian Huang, Lingpeng Kong, Jiajun Chen, and Lei Li. 2023.
\newblock \href {https://arxiv.org/pdf/2304.04675} {Multilingual machine translation with large language models: Empirical results and analysis}.
\newblock \emph{arXiv preprint arXiv:2304.04675}.

\end{thebibliography}

\clearpage

\appendix

\section*{Appendix}
In this Appendix, we first provide additional implementation details in Section~\ref{sec:more_imple_detail}. Then, Section~\ref{sec:ds_stats} provides more details about the sign language dataset used in this study, including its statistics. 
In Section~\ref{sec:add_abl_res}, we present further experimental results. Finally, in Section~\ref{sec:appen_lvlms}, we discuss the feasibility of existing Vision-Language Models (VLMs) in the SLT domain.

\section{More Implementation Details}\label{sec:more_imple_detail}

\subsection{Components of SpaMo.}

In the SA module, we utilize two distinct linear projection layers tailored for the output feature of ME and SE. For short-term modeling, we employ a 1D TCN configured with a specific sequence of layers: \(\{K5, P2, K5, P2\}\), where \(K_\sigma\) represents a kernel size of \(\sigma\), and \(P_\sigma\) indicates a pooling layer with a kernel size of \(\sigma\)~\cite{hu2023continuous}. To integrate features into the LLM's embedding space, we leverage an MLP cross-modal connector~\cite{liu2024improved}, projecting the features into a 2048-dimensional space. 

\subsection{Prompt Template.}

To focus the LLM on the SLT task, we employ a specific prompting strategy. Our prompt includes a clear instructive prompt: ``Translate the given sentence into German.'' Following this, we incorporate multilingual translations via a translation engine such as Google Translate\footnote{\scriptsize\texttt{\url{https://cloud.google.com/translate}}}, which are sampled from the training set. These translations are included to facilitate In-Context Learning (ICL)~\cite{brown2020language}. The prompt is structured as follows: ``Translate the given sentence into German. [SRC] = [TRG].'' Here, the source input (e.g., a sentence in French) serves as the foreign language example, and the corresponding response is the translation into the target language (e.g., German, as used in PHOENIX14T). 
An example of this prompt structure is provided in Table~\ref{tab:prompt}. To ensure that the LLM does not directly access the target translations during training, we shuffle the translation samples so that they do not match the target translation. At test time, we select a translation pair from the training set to use as a reference.

\begin{table}[t]
    \scriptsize
    \centering
    \resizebox{\linewidth}{!}{%
        \renewcommand{\arraystretch}{0.95}
        \begin{tabular} {rl} \toprule
            Sign Video Input:& [Extracted Sign Feature] \\ 
            \noalign{\vskip 0.3ex}\cdashline{1-2}\noalign{\vskip 0.7ex}
            Instruction:& Translate the given sentence into German. \\
            \noalign{\vskip 0.3ex}\cdashline{1-2}\noalign{\vskip 0.7ex}
            \multirow{11}{*}{In Context Examplars:}& Soil frost is possible there and in the southern \\& low mountain ranges.=dort sowie in den südlichen \\& mittelgebirgen ist bodenfrost möglich. \\& \\& La helada del suelo es posible allí y en las cadenas \\& montañosas del sur.=dort sowie in den südlichen \\& mittelgebirgen ist bodenfrost möglich. \\& \\&Le gel du sol est possible là-bas et dans les chaînes \\& de montagnes basses du sud.=dort sowie in den südlichen \\& mittelgebirgen ist bodenfrost möglich. \\ \bottomrule
        \end{tabular}
    }
    \caption{An example of prompt used in this paper.}
    \label{tab:prompt}
\end{table}
\begin{table}[t!]
\scriptsize
\centering
\resizebox{\linewidth}{!}{%
    \renewcommand{\arraystretch}{0.95}
    \begin{tabular} {l ccccc} \toprule
    \textbf{Visual Encoders} (SE + ME)& \textbf{B1} & \textbf{B2} & \textbf{B3} & \textbf{B4} & \textbf{RG} \\ \midrule
    DINOv2 + V-JEPA& 45.67& 32.94& 25.27& 20.35& 41.32 \\
    DINOv2 + VideoMAE& 47.31& 34.60& 26.90& 21.86& 42.50 \\
    CLIP + V-JEPA& 47.82& 34.71& 26.76& 21.66& 43.68 \\ \noalign{\vskip 0.3ex}\cdashline{1-6}\noalign{\vskip 0.7ex}
    CLIP + VideoMAE& \textbf{49.80}& \textbf{37.32} &\textbf{29.50} &\textbf{24.32} &\textbf{46.57} \\ \bottomrule
    \end{tabular}
}
\caption{Ablation study on various combinations of visual encoders. The results are with VT-Align.}
\label{tab:abl_vis_enc}
\end{table}
\begin{table}[t!]
\scriptsize
\centering
\resizebox{\linewidth}{!}{%
    \renewcommand{\arraystretch}{0.95}
    \begin{tabular} {l ccccc} \toprule
   \textbf{ Methods}& \textbf{B1}& \textbf{B2}& \textbf{B3}& \textbf{B4}& \textbf{RG} \\ \midrule
    Ours (w/o LoRA)& 46.11& 32.65& 24.69& 19.67& 42.91 \\
    Ours (w/ LoRA)& \textbf{49.80}& \textbf{37.32} &\textbf{29.50} &\textbf{24.32} &\textbf{46.57} \\ \bottomrule
    \end{tabular}
}
\caption{Ablation on our method with and without LoRA.}
\label{tab:abl_rola}
\end{table}
\begin{table*}[t!]
    \centering
    \scriptsize
    \resizebox{\linewidth}{!}{%
        \renewcommand{\arraystretch}{0.95}
        \begin{tabular}{l c c c c c l} 
        \toprule 
         \textbf{Dataset} &\textbf{Language} &\textbf{\#Vocab} &\textbf{Train} / \textbf{Valid} / \textbf{Test}& \textbf{Avg. No. Frame}& \textbf{Gloss}& \textbf{Domain}\\ \midrule
         PHOENIX14T~\cite{camgoz2018neural}&   DGS&3K& 7,096 / 519 / 642&  116 & \(\checkmark\)& Weather Forecast \\
         CSL-Daily~\cite{zhou2021improving}&   CSL&2K& 18,401 / 1,077 / 1,176&  119& \(\checkmark\)& Daily-life \\
         How2Sign~\cite{duarte2021how2sign} &   ASL&16K& 31,128 / 1,741 / 2,322&  173& \xmark& Instructional \\ \bottomrule
        \end{tabular}
    }
    \caption{Statistics of three sign language datasets used in this work. DGS: German Sign Language; CSL: Chinese Sign Language; ASL: American Sign Language; Avg. No. Frame: average number of video frames.} 
    \label{tab:dataset_stats}
\end{table*}

\subsection{Training.}
For training, we use the AdamW optimizer~\cite{loshchilov2017decoupled}, with \(\beta_1=0.9\), \(\beta_2=0.98\), and a weight decay of \(0.01\). The learning rate schedule includes a cosine decay with a peak learning rate of 1e-4 and a linear warmup of over 10K steps, with a minimum learning rate of 5e-5. We train our model for 40 epochs, using a single NVIDIA A100 GPU, completing the entire process within 24 hours.

\subsection{Evaluating Process with KDEs.}\label{sec:appen_kde}
To evaluate the quality of the learned representations, we utilize Kernel Density Estimation (KDE) to estimate the probability density functions of the embeddings from GFSLT-VLP and ours. 
Due to different dimensionality between these methods (1,024 vs. 2,048), we run Principal Component Analysis (PCA) to reduce the number of dimensions while retaining the most significant variance components. 
This dimensionality reduction facilitates more efficient and stable KDE fitting. KDE can be expressed as:
\begin{equation}
    f_\text{kde}(\mathbf{z}) = \frac{1}{n h^d} \sum_{i=1}^{n} K\left(\frac{\mathbf{z} - \mathbf{z_i}}{h}\right),
\end{equation}
where \(\mathbf{z_i}\) denotes the representation points, \( K \) denotes the kernel function, \( h \) is the bandwidth parameter, \( d \) is the dimensionality of the data, and \( n \) is the number of data points.

The entropy of KDE is then calculated as:
\begin{equation}
    H = -\sum_{i=1}^{n} f_{\text{kde}}(\mathbf{z_i}) \log f_{\text{kde}}(\mathbf{z_i}),
\end{equation}
where \( H \) represents the entropy, and \( f(\mathbf{z_i}) \) are the estimated density values at the representation points. 

\subsection{Generating Visual Tokens}\label{sec:append_gen_vis_tok}

Inspired by the reverse engineering~\cite{ju2023continuous}, we first compute the Euclidean distance between the sign feature \( Z_{sm} \) and the LLM's embedding table \( E_{llm} \in \mathbb{R}^{V \times d'} \), where \( V \) represents the vocabulary size. Each sign feature is then mapped to the word associated with the shortest distance in this space. This process can be expressed as \(\text{dist}(Z_{sm}, E_{llm}) \leq \Delta\), where \( \text{dist}(\cdot) \) denotes the Euclidean distance function, and \( \Delta \) represents the shortest distance to \( E_{llm} \) across all sign features.

\section{Statistics of Sign Language Datasets}\label{sec:ds_stats}

Table~\ref{tab:dataset_stats} presents a comparative overview of three popular sign language datasets: PHOENIX14T, CSL-Daily, and How2Sign, each with distinct statistics and domain.

PHOENIX14T focuses on German Sign Language (DGS) within the specific domain of weather forecasting, featuring a relatively small vocabulary of 3K words and a concise average video length of 116 frames. It includes 7,096 training samples, 519 validation samples, and 642 test samples, with gloss annotations available. This dataset is tailored for domain-specific tasks, offering clear and repetitive patterns ideal for translation and recognition within weather-related contexts.

In comparison, CLS-Daily, a dataset for Chinese Sign Language (CSL), covers a broader range of topics than PHOENIX14T, spanning areas such as family life, medical care, school, banking, shopping, and social interactions. It features a vocabulary of 2K words and an average video length of 119 frames. The dataset includes 18,401 training samples, 1,077 validation samples, and 1,176 test samples, also with gloss annotations.

On the other hand, How2Sign focuses on American Sign Language (ASL) in the instructional domain. It offers a significantly larger and more diverse dataset, with a vocabulary of 16K words and an average video length of 173 frames. The dataset consists of 31,128 training samples, 1,741 validation samples, and 2,322 test samples, but lacks gloss annotations. The diversity and complexity of How2Sign make it particularly suitable for general sign language related tasks, especially those that involve understanding varied and intricate sign sequences.

\section{More Experiments}\label{sec:add_abl_res}

\subsection{Effect of Visual Encoders.}
We assess the effect of various combinations of visual encoders (SE \& ME). Table~\ref{tab:abl_vis_enc} shows four different encoders: DINOv2~\cite{oquab2023dinov2}, CLIP~\cite{radford2021learning}, V-JEPA~\cite{bardes2024revisiting}, and VideoMAE~\cite{tong2022videomae}. The results demonstrate that the combination of CLIP and VideoMAE delivers the highest performance, suggesting potential for further improvement as visual encoders continue to advance.

\subsection{Effect of LoRA.}
We evaluate the effect of LoRA on the LLM. As illustrated in Table~\ref{tab:abl_rola}, the LLM with LoRA demonstrates superior performance.

\begin{table}[t!]
\scriptsize
\centering
\resizebox{\linewidth}{!}{%
    \renewcommand{\arraystretch}{0.95}
    \begin{tabular} {l cccc c} \toprule
    \textbf{Methods}& \textbf{Vis. Ft.}& \textbf{\#Trainable Params}& \textbf{\#Total Params}& \textbf{B4} \\ \midrule
    GFSLT-VLP~\cite{zhou2023gloss}& \(\checkmark\)& 215.6M& 215.6M& 21.44 \\
    Sign2GPT~\cite{wong2024sign2gpt}& \(\checkmark\)& 16M& 1.8B& 22.52 \\
    Fla-LLM~\cite{chen2024factorized}& \(\checkmark\)& >705.6M*& >705.6M*& 23.09 \\
    SignLLM~\cite{gong2024llms}& \(\checkmark\)& -& >7B*& 23.40 \\
    \noalign{\vskip 0.3ex}\cdashline{1-5}\noalign{\vskip 0.7ex}
    \textbf{SpaMo (Ours)}& \xmark& 22.7M& 3.5B& \textbf{24.32} \\ \bottomrule
    \end{tabular}
    }
\caption{Model parameter comparison. * denotes an estimated value due to the unavailability of public code. ``Vis. Ft.'' denotes to the visually fine-tuned on sign language datasets.}
\label{tab:params}
\end{table}

\subsection{Parameter Comparisons}

We present a comparison of various SLT methods, focusing on the presence of visual fine-tuning ("Vis. Ft."), the number of trainable and total parameters, and their performance as measured by the BLEU-4 score.
As shown in Table~\ref{tab:params}, our method, SpaMo, achieves the highest BLEU-4 score of 24.32 without the need for the visual fine-tuning. 
SpaMo requires 22.7M trainable parameters, which is relatively efficient compared to other methods such as GFSLT-VLP (215.6M), Sign2GPT (16M), and Fla-LLM (>705.6M). 
This demonstrates that SpaMo effectively balances model complexity and training efficiency to achieve superior performance without the additional step of the visual fine-tuning.

\subsection{More Qualitative Results}\label{sec:add_qual_res}

We provide additional translation examples for PHOENIX14T, CSL-Daily, and How2Sign. 
As shown in Table~\ref{tab:qaul_res_p14t}, in PHOENIX14T, our method consistently delivers accurate translations, while GFSLT-VLP struggles to capture the correct semantic meaning. 

In CSL-Daily, we present a comparison between glosses and visual tokens, as well as between reference translations and generated translations. 
As shown in Table~\ref{tab:qaul_res_csl}, most visual tokens are matched to the glosses, though they are not perfectly aligned. 
Notably, in the last three examples, the visual tokens include words that are missing from the glosses but appear in the reference translations.

For How2Sign, Table~\ref{tab:qaul_res_h2s} presents translation results along with their corresponding visual tokens. Since How2Sign lacks gloss annotations, we include actual sign frames for qualitative comparison. 
Similar to the results on PHOENIX14T and CSL-Daily, many visual tokens in How2Sign are closely aligned with the translations. 
Note that although OpenSLT~\cite{tarres2023sign} is the only publicly available baseline\footnote{\scriptsize\texttt{\url{https://github.com/imatge-upc/slt_how2sign_wicv2023}}}, we were unable to reproduce their results due to a broken link to the fine-tuned I3D features at the time of drafting.

\begin{table}[t]
\scriptsize
\centering
\resizebox{\linewidth}{!}{%
    \renewcommand{\arraystretch}{0.95}
    \begin{tabular} {l c} \toprule
    \textbf{Pairs}& \textbf{Cosine Similarity} \\ \toprule
    Glosses \& Target Translations&  0.8400 \\ 
    Visual Tokens \& Target Translations& 0.6781 \\ 
    \noalign{\vskip 0.3ex}\cdashline{1-2}\noalign{\vskip 0.7ex}
    Visual Tokens \& Glosses& 0.6779 \\ \bottomrule
    \end{tabular}
}
\caption{Cosine similarity between the visual tokens and the glosses.}
\label{tab:vis_tok_sim}
\end{table}

\begin{figure}[t]
    \centering
    \includegraphics[trim=0cm 0cm 0cm 0cm,clip=true,width=\linewidth]{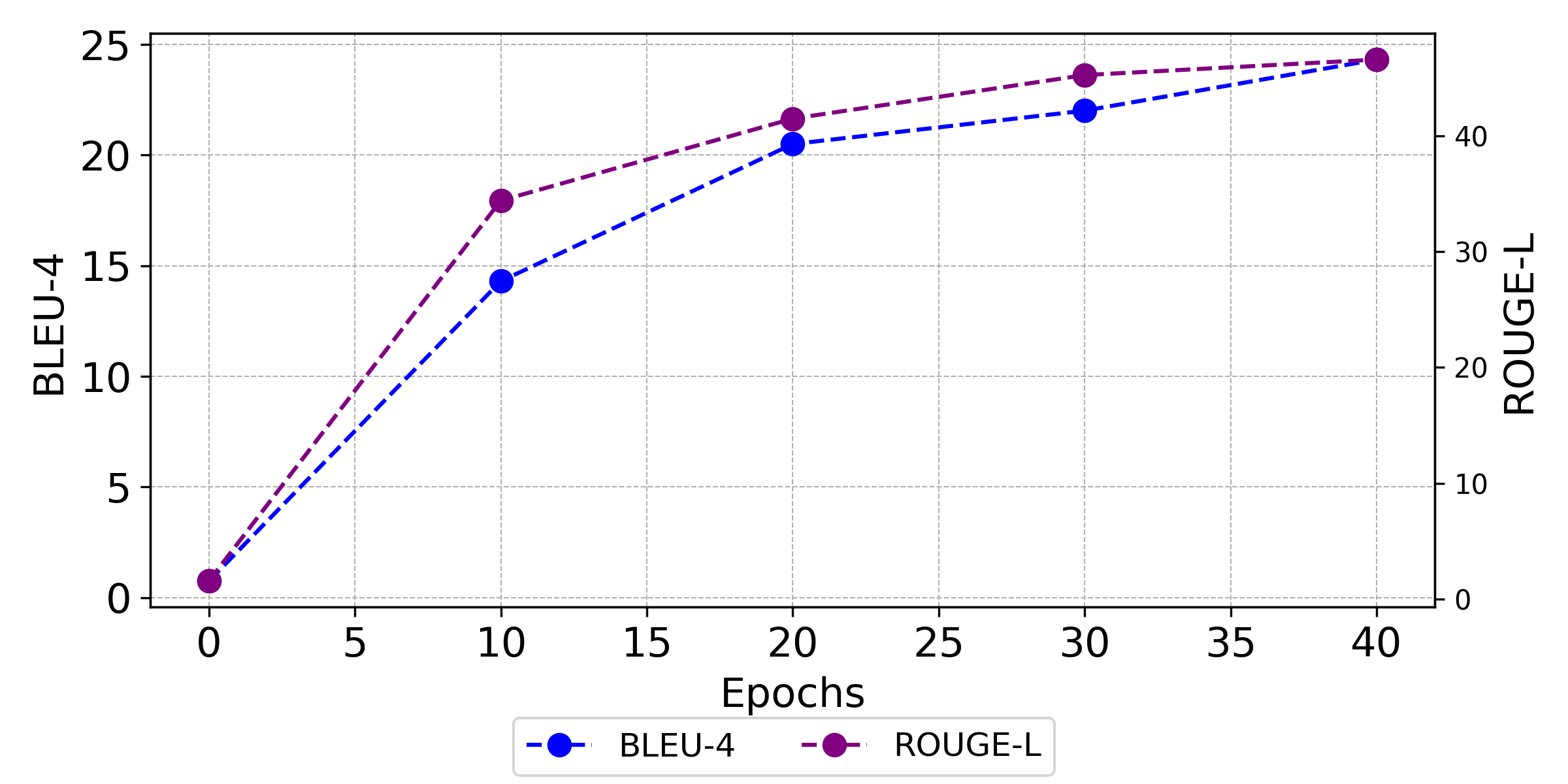}
    \caption{Performance curves across epochs.} 
    \label{fig:p_curve_epochs}
    \vspace{-1em}
\end{figure}

\subsection{Cosine Similarity Between Visual Tokens and Glosses}

We use Sentence-BERT~\cite{reimers2019sentence} to evaluate the similarity between the generated visual tokens and the ground-truth glosses from PHOENIX14T, using cosine similarity. Additionally, we assess the similarity between the visual tokens and the target translation, as well as the alignment between the glosses and the translation, which highlights varying degrees of correspondence.

As shown in Table~\ref{tab:vis_tok_sim}, and as we expected, the highest similarity occurs between glosses and target translations, indicating a strong semantic correspondence. By contrast, the visual tokens show lower similarity to both target translations and glosses, suggesting that they are not random. The reason for the score not being higher is likely the inclusion of unrelated words in the visual tokens compared to the actual glosses, as illustrated in Table~\ref{tab:vistok_res_p14t}.

\subsection{Performance Curve Across Epochs}

Figure~\ref{fig:p_curve_epochs} shows the performance curves of SpaMo using BLEU-4 and ROUGE over 40 training epochs on the PHOENIX14T dataset. For comparison, we note that other models such as SignLLM, Sign2GPT, and Fla-LLM are trained for 20, 100, and 75 epochs, respectively. These results highlight the progressive improvements in SpaMo's performance as training advances, offering a detailed look at its efficiency relative to other models.

\section{Feasibility of Existing VLMs in SLT}\label{sec:appen_lvlms}

Recent advancements in Vision-Language Models (VLMs)~\cite{alayrac2022flamingo,li2023blip,liu2024visual,cheng2024videollama} have enabled LLMs to comprehend various modalities including images and videos, beyond just text. However, in the SLT domain, current VLM designs are not well-suited for processing long sequences of sign videos. As shown in Table~\ref{tab:dataset_stats}, the average sign video length exceeds 116 frames, which is significantly longer than typical action recognition or video-text datasets, where sample lengths are often under 16 frames. 
For example, Flamingo~\cite{alayrac2022flamingo}, a widely recognized vision-language model, uses 8 frames during training and only 32 frames during inference—far fewer than what is required for SLT. Similarly, VideoLlama2~\cite{cheng2024videollama} employs 8 frames for training. Moreover, recent LLM-based SLT methods~\cite{wong2024sign2gpt,gong2024llms}, including our method, can be classified within the VLM family but are specialized to process and capture long sign video sequences.

\begin{table*}[t]
\scriptsize
\centering
\resizebox{\linewidth}{!}{%
    \begin{tabular} {rl} \toprule
        \multirow{2}{*}{Ref:}& und nun die wettervorhersage für morgen sonntag den zwölften juli. \\
                & \textit{(and now the weather forecast for tomorrow Sunday the twelfth of July.)} \\
        \multirow{2}{*}{GFSLT-VLP:}& \textcolor{blue}{und nun die wettervorhersage für morgen sonntag den zwölften} \textcolor{red}{juni}. \\
                            & \textit{(and now the weather forecast for tomorrow, Sunday, the twelfth of June.)} \\
        \multirow{2}{*}{Ours:}& \textcolor{blue}{und nun die wettervorhersage für morgen sonntag den zwölften juli}. \\
                            & \textit{(and now the weather forecast for tomorrow Sunday the twelfth of July.)} \\ \midrule
                            
        \multirow{2}{*}{Ref:}& in der nacht muss vor allem in der nordwesthälfte mit schauern und gewittern gerechnet werden die heftig ausfallen können. \\
                            & \textit{(During the night, showers and thunderstorms are expected, especially in the northwest half, which could be heavy.)} \\
        \multirow{2}{*}{GFSLT-VLP:}& heute \textcolor{blue}{nacht} gibt es im \textcolor{blue}{norden} teilweise \textcolor{blue}{kräftige} \textcolor{blue}{schauer} und \textcolor{blue}{gewitter die} örtlich unwetterartig sein \textcolor{blue}{können}. \\
                            & \textit{(At night, showers and thunderstorms can be expected, especially in the northwest half, which can sometimes be strong.)} \\
        \multirow{2}{*}{Ours:}& \textcolor{blue}{in der nacht muss vor allem in der nordwesthälfte mit schauern und gewittern gerechnet werden die} mitunter kräftig sein \textcolor{blue}{können}. \\
                            & \textit{(During the night, showers and thunderstorms are expected, particularly in the northwest half, which may be heavy.)} \\ \midrule

        \multirow{2}{*}{Ref:}& und nun die wettervorhersage für morgen donnerstag den siebenundzwanzigsten august. \\
                            & \textit{(and now the weather forecast for tomorrow, Thursday the twenty-seventh of August.)} \\
        \multirow{2}{*}{GFSLT-VLP:}& \textcolor{blue}{und nun die wettervorhersage für morgen donnerstag den} \textcolor{red}{sechsundzwanzigsten} \textcolor{blue}{august}. \\
                            & \textit{(and now the weather forecast for tomorrow, Thursday the twenty-sixth of August.)} \\
        \multirow{2}{*}{Ours:}& \textcolor{blue}{und nun die wettervorhersage für morgen donnerstag den siebenundzwanzigsten august}.  \\
                            & \textit{(and now the weather forecast for tomorrow, Thursday the twenty-seventh of August.)} \\ \midrule

        \multirow{2}{*}{Ref:}& am tag ist es im westen freundlich sonst sonne und dichtere wolken im wechsel hier und da fallen einzelne schauer. \\
                            & \textit{(During the day it is friendly in the west, otherwise sun and denser clouds alternate, with occasional showers here and there)} \\
        \multirow{2}{*}{GFSLT-VLP:}& \textcolor{blue}{am tag wechseln} \textcolor{blue}{sonne und} \textcolor{blue}{wolken} einander ab \textcolor{blue}{im westen fallen} mitunter gewittrige \textcolor{blue}{schauer}.  \\
                            & \textit{(During the day sun and clouds alternate, in the west, occasional stormy showers may occur)} \\
        \multirow{2}{*}{Ours:}& \textcolor{blue}{am} \textcolor{blue}{tag} \textcolor{blue}{ist} \textcolor{blue}{es} \textcolor{blue}{im} \textcolor{blue}{westen} \textcolor{blue}{freundlich} mit \textcolor{blue}{sonne und} dichteren \textcolor{blue}{wolken hier und da fallen schauer}. \\
                            & \textit{(During the day it is friendly in the west with sun and denser clouds, with occasional showers here and there)} \\ \midrule

        \multirow{2}{*}{Ref:}& abseits der gewittern weht der wind schwach bis mäßig an der küste frisch. \\
                            & \textit{(Away from the thunderstorms, the wind blows weak to moderate, fresh at the coast.)} \\
        \multirow{2}{*}{GFSLT-VLP:}& abgesehen von gewitterböen schwacher \textcolor{blue}{bis} mäßiger \textcolor{blue}{an} den küsten auch frischer wind \\
                            & \textit{(Apart from thunderstorm gusts, weak to moderate, also fresh wind at the coasts.)} \\
        \multirow{2}{*}{Ours:}& \textcolor{blue}{abseits} \textcolor{blue}{der} \textcolor{blue}{gewittern} \textcolor{blue}{weht} \textcolor{blue}{der} \textcolor{blue}{wind} \textcolor{blue}{schwach bis} \textcolor{blue}{mäßig an} den küsten auch \textcolor{blue}{frisch}. \\
                            & \textit{(Away from the thunderstorms, the wind blows weak to moderate, also fresh at the coasts.)} \\ \midrule


        \multirow{2}{*}{Ref:}& am sonntag im norden und an den alpen mal sonne mal wolken und ab und an schauer sonst ist es recht freundlich. \\
                            & \textit{(On Sunday in the north and in the Alps sometimes sun sometimes clouds and occasional showers otherwise it is quite pleasant.)} \\
        \multirow{2}{*}{GFSLT-VLP:}& \textcolor{blue}{am} \textcolor{blue}{sonntag} \textcolor{blue}{im} \textcolor{blue}{norden an} \textcolor{blue}{den} \textcolor{blue}{alpen} einige \textcolor{blue}{schauer} \textcolor{blue}{sonst} \textcolor{blue}{ist} \textcolor{blue}{es} \textcolor{blue}{recht} \textcolor{blue}{freundlich}.  \\
                            & \textit{(On Sunday in the north in the Alps some showers otherwise it is quite pleasant.)} \\
        \multirow{2}{*}{Ours:}& \textcolor{blue}{am} \textcolor{blue}{sonntag} \textcolor{blue}{im} \textcolor{blue}{norden} \textcolor{blue}{und} \textcolor{blue}{an} \textcolor{blue}{den} \textcolor{blue}{alpen} \textcolor{blue}{mal} \textcolor{blue}{sonne} \textcolor{blue}{mal} \textcolor{blue}{wolken} \textcolor{blue}{und} nur einzelne \textcolor{blue}{schauer} \textcolor{blue}{sonst} meist \textcolor{blue}{freundlich}. \\
                            & \textit{(On Sunday in the north and in the Alps sometimes sun sometimes clouds and only a few showers otherwise mostly pleasant.)} \\ \midrule

        \multirow{2}{*}{Ref:}& am mittwoch eine mischung aus sonne wolken und nebelfeldern im nordwesten hier und da schauer sonst ist es trocken. \\
                            & \textit{(On Wednesday a mix of sun, clouds, and fog patches in the northwest; here and there showers, otherwise it is dry.)} \\
        \multirow{2}{*}{GFSLT-VLP:}& \textcolor{blue}{am} \textcolor{blue}{mittwoch} gibt \textcolor{blue}{es} viele \textcolor{blue}{wolken} \textcolor{blue}{hier und} \textcolor{blue}{da} \textcolor{blue}{schauer} vor allem \textcolor{blue}{im} \textcolor{blue}{nordwesten} bleibt \textcolor{blue}{es} meist \textcolor{blue}{trocken}. \\
                            & \textit{(On Wednesday there will be many clouds; here and there showers, especially in the northwest, it remains mostly dry.)} \\
        \multirow{2}{*}{Ours:}& \textcolor{blue}{am} \textcolor{blue}{mittwoch} \textcolor{blue}{eine} \textcolor{blue}{mischung} \textcolor{blue}{aus} \textcolor{blue}{sonne} \textcolor{blue}{wolken und} nebel \textcolor{blue}{im} \textcolor{blue}{nordwesten} einige \textcolor{blue}{schauer sonst} bleibt \textcolor{blue}{es} meist \textcolor{blue}{trocken}. \\
                            & \textit{(On Wednesday a mix of sun, clouds, and fog in the northwest; some showers, otherwise it remains mostly dry.)} \\ \midrule

        \multirow{2}{*}{Ref:}& am tag scheint verbreitet die sonne im süden und westen bilden sich später gebietsweise quellwolken. \\
                            & \textit{(During the day, the sun shines widely in the south, and later, isolated cumulus clouds form in the west.)} \\
        \multirow{2}{*}{GFSLT-VLP:}& \textcolor{blue}{am} \textcolor{blue}{tag} \textcolor{blue}{scheint} in der südhälfte häufig \textcolor{blue}{die} \textcolor{blue}{sonne} hier \textcolor{blue}{und} da ein paar wolken. \\
                            & \textit{(During the day, the sun often shines in the southern half, here and there a few clouds.)} \\
        \multirow{2}{*}{Ours:}& \textcolor{blue}{am} \textcolor{blue}{tag} \textcolor{blue}{scheint} \textcolor{blue}{verbreitet} \textcolor{blue}{die} \textcolor{blue}{sonne} \textcolor{blue}{im} \textcolor{blue}{süden und} im äußersten \textcolor{blue}{westen} tauchen hier \textcolor{blue}{und} da ein paar \textcolor{blue}{quellwolken} auf. \\
                            & \textit{(During the day, the sun shines widely in the south, and in the far west, here and there, a few cumulus clouds appear.)} \\ \midrule

        \multirow{2}{*}{Ref:}& der wind weht mäßig bis frisch mit starken bis stürmischen böen im bergland teilweise schwere sturmböen im südosten mitunter nur schwacher wind. \\
                            & \textit{(The wind blows moderately to freshly with strong to stormy gusts in the mountainous regions, partly severe storm gusts in the southeast, occasionally only weak wind.)} \\
        \multirow{2}{*}{GFSLT-VLP:}& \textcolor{blue}{der} \textcolor{blue}{wind} \textcolor{blue}{weht} \textcolor{blue}{mäßig} \textcolor{blue}{bis} \textcolor{blue}{frisch} bei schauern sowie \textcolor{blue}{im} \textcolor{blue}{südosten} \textcolor{blue}{schwere sturmböen} \textcolor{blue}{im} \textcolor{blue}{bergland} starker \textcolor{blue}{bis} stürmböen. \\
                            & \textit{(The wind blows moderately to freshly with showers, as well as severe storm gusts in the southeast, in the mountainous regions strong to stormy gusts.)} \\
        \multirow{2}{*}{Ours:}& \textcolor{blue}{der} \textcolor{blue}{wind} \textcolor{blue}{weht} \textcolor{blue}{mäßig} \textcolor{blue}{bis} \textcolor{blue}{frisch} \textcolor{blue}{mit} \textcolor{blue}{starken} \textcolor{blue}{bis} \textcolor{blue}{stürmischen} \textcolor{blue}{böen} \textcolor{blue}{auf} \textcolor{blue}{den} \textcolor{blue}{bergen} \textcolor{blue}{schwere sturmböen} \textcolor{blue}{im} süden \textcolor{blue}{sonst} \textcolor{blue}{schwacher} \textcolor{blue}{wind}. \\
                            & \textit{(The wind blows moderately to freshly with strong to stormy gusts on the mountains, severe storm gusts in the south, otherwise weak wind.)} \\ \midrule

        \multirow{2}{*}{Ref:}& am montag überall wechselhaft und deutlich kühler. \\
                            & \textit{(On Monday, everywhere is changeable and significantly cooler.)} \\
        \multirow{2}{*}{GFSLT-VLP:}& \textcolor{blue}{am} \textcolor{blue}{montag} wird es wieder \textcolor{blue}{wechselhafter} \textcolor{blue}{kühler}. \\
                                    & \textit{(On Monday, it will be changeable and cooler again.)} \\
        \multirow{2}{*}{Ours:}& \textcolor{blue}{am} \textcolor{blue}{montag} \textcolor{blue}{überall} \textcolor{blue}{wechselhaft und}  \textcolor{blue}{deutlich} \textcolor{blue}{kühler}. \\
                                    & \textit{(On Monday, everywhere is changeable and significantly cooler.)} \\ \midrule

        \multirow{2}{*}{Ref:}& sonst ein wechsel aus sonne und wolken. \\
                & \textit{(Otherwise a mix of sun and clouds.)} \\
        \multirow{2}{*}{GFSLT-VLP:}& ansonsten wechseln sich \textcolor{red}{teilweise} dichte \textcolor{blue}{wolken und} \textcolor{blue}{sonne} ab. \\
                            & \textit{(Otherwise partially dense clouds and sun alternate.)} \\
        \multirow{2}{*}{Ours:}& \textcolor{blue}{sonst} \textcolor{blue}{ein} \textcolor{blue}{wechsel} \textcolor{blue}{aus} \textcolor{blue}{sonne und} \textcolor{blue}{wolken}. \\
                            & \textit{(Otherwise a mix of sun and clouds.)} \\ \midrule

        \multirow{2}{*}{Ref:}& und nun die wettervorhersage für morgen samstag den sechsundzwanzigsten januar. \\
                            & \textit{And now the weather forecast for tomorrow, Saturday, the twenty-sixth of January.} \\
        \multirow{2}{*}{GFSLT-VLP:}& \textcolor{blue}{und} \textcolor{blue}{nun} \textcolor{blue}{die} \textcolor{blue}{wettervorhersage} \textcolor{blue}{für} \textcolor{blue}{morgen} \textcolor{blue}{samstag} \textcolor{blue}{den} \textcolor{blue}{sechsundzwanzigsten} \textcolor{red}{dezember}. \\
                            & \textit{And now the weather forecast for tomorrow, Saturday, the twenty-sixth of December.} \\
        \multirow{2}{*}{Ours:}& \textcolor{blue}{und} \textcolor{blue}{nun} \textcolor{blue}{die} \textcolor{blue}{wettervorhersage} \textcolor{blue}{für} \textcolor{blue}{morgen} \textcolor{blue}{samstag} \textcolor{blue}{den} \textcolor{blue}{sechsundzwanzigsten} \textcolor{blue}{januar}. \\
                            & \textit{And now the weather forecast for tomorrow, Saturday, the twenty-sixth of January.} \\ \midrule

        \multirow{2}{*}{Ref:}& sonst ist es recht freundlich. \\
                            & \textit{Otherwise it is quite pleasant.} \\
        \multirow{2}{*}{GFSLT-VLP:}& \textcolor{blue}{sonst} \textcolor{red}{überwiegend} \textcolor{blue}{freundlich}. \\
                            & \textit{Otherwise mostly pleasant.} \\
        \multirow{2}{*}{Ours:}& \textcolor{blue}{sonst} \textcolor{blue}{ist} \textcolor{blue}{es} \textcolor{blue}{recht} \textcolor{blue}{freundlich}. \\
                            & \textit{Otherwise it is quite pleasant.} \\ \bottomrule
    \end{tabular}
}
\caption{Translation results on the test set compared to GFSLT-VLP on PHOENIX14T. Correctly translated 1-grams are highlighted in \textcolor{blue}{blue}, while incorrect translations are marked in \textcolor{red}{red}.}
\label{tab:qaul_res_p14t}
\end{table*}
\AtBeginDvi{\input{zhwinfonts}}
\begin{CJK*}{UTF8}{gbsn}
\begin{table*}[t]
\scriptsize
\centering
\resizebox{0.8\linewidth}{!}{%
    \begin{tabular} {rl} \toprule 
        Gloss:& 你\hspace{1mm}小\hspace{1mm}张\hspace{1mm}什么\hspace{1mm}时间\hspace{1mm}认识 \\ 
        Vis. Token:& \colorbox{green!40}{你}\hspace{1mm}\colorbox{green!40}{小}三 \colorbox{green!40}{张}三 机场 哪里 \colorbox{green!40}{什么} \colorbox{green!40}{时候} 桌 \colorbox{green!40}{认识} \\
        \noalign{\vskip 0.3ex}\cdashline{1-2}\noalign{\vskip 0.7ex}
        Ref:& 你\hspace{1mm}和\hspace{1mm}小张\hspace{1mm}什么\hspace{1mm}时候\hspace{1mm}认识\hspace{1mm}的? \\
        Ours:& \textcolor{blue}{你\hspace{1mm}和\hspace{1mm}小张\hspace{1mm}什么\hspace{1mm}时候\hspace{1mm}认识\hspace{1mm}的?} \\ \midrule 

        Gloss:& 椅子\hspace{1mm}他们\hspace{1mm}想\hspace{1mm}什么\hspace{1mm}时间\hspace{1mm}去\hspace{1mm}买 \\
        Vis. Token:& \colorbox{green!40}{椅子} 时候 什么 你们 \colorbox{green!40}{他们} 他 \colorbox{green!40}{想} 每天 \colorbox{green!40}{什么} \colorbox{green!40}{时候} 旅游 \colorbox{green!40}{去} 女儿 \colorbox{green!40}{买} 考试? \\
        \noalign{\vskip 0.3ex}\cdashline{1-2}\noalign{\vskip 0.7ex}
        Ref:& 他们\hspace{1mm}想\hspace{1mm}什么\hspace{1mm}时候\hspace{1mm}去\hspace{1mm}买\hspace{1mm}椅子? \\
        Ours:& \textcolor{blue}{他们\hspace{1mm}想\hspace{1mm}什么\hspace{1mm}时候\hspace{1mm}去\hspace{1mm}买\hspace{1mm}椅子?} \\ \midrule 

        Gloss:& 不是\hspace{1mm}他\hspace{1mm}去\hspace{1mm}见面\hspace{1mm}同学 \\
        Vis. Token:& \colorbox{green!40}{不是}为什么一个\colorbox{green!40}{他} \colorbox{green!40}{去} 看见他人看\colorbox{green!40}{同学}桌？ \\
        \noalign{\vskip 0.3ex}\cdashline{1-2}\noalign{\vskip 0.7ex}
        Ref:& 不是,\hspace{1mm}他\hspace{1mm}是\hspace{1mm}去\hspace{1mm}见 他\hspace{1mm}同学 。 \\
        Ours:& \textcolor{blue}{不是,\hspace{1mm}他\hspace{1mm}是\hspace{1mm}去}\hspace{1mm}看\hspace{1mm}\textcolor{blue}{同学 。} \\ \midrule 

        Gloss:& 这\hspace{1mm}衣服\hspace{1mm}红\hspace{1mm}怎么样\hspace{1mm}这\hspace{1mm}是\hspace{1mm}新 \\
        Vis. Token:& 孩子 北京\colorbox{green!40}{衣服}\hspace{1mm}\colorbox{green!40}{红色}喜欢我他\colorbox{green!40}{怎么样}认识跑身体跑 \colorbox{green!40}{这} 个他 \colorbox{green!40}{新} 北京多? \\
        \noalign{\vskip 0.3ex}\cdashline{1-2}\noalign{\vskip 0.7ex}
        Ref:& 这\hspace{1mm}件\hspace{1mm}红色\hspace{1mm}的\hspace{1mm}衣服\hspace{1mm}怎么样?\hspace{1mm}这是\hspace{1mm}新\hspace{1mm}的 。 \\
        Ours:& \textcolor{blue}{这\hspace{1mm}件\hspace{1mm}红色\hspace{1mm}的\hspace{1mm}衣服\hspace{1mm}怎么样?}\hspace{1mm}是\hspace{1mm}今年 \hspace{1mm}\textcolor{blue}{的 。} \\ \midrule 

        Gloss:& 中午\hspace{1mm}吃\hspace{1mm}好了\hspace{1mm}要\hspace{1mm}多\hspace{1mm}吃\hspace{1mm}水果 \\
        Vis. Token:& 饮料 \colorbox{green!40}{吃} 饭 \colorbox{green!40}{吃} 了冷了 \colorbox{green!40}{完} 了老板安全 \colorbox{green!40}{要} 粥早\colorbox{green!40}{多} 喜欢饮料吃饭 \colorbox{green!40}{水果} 吃饭 \\
        \noalign{\vskip 0.3ex}\cdashline{1-2}\noalign{\vskip 0.7ex}
        Ref:& 吃\hspace{1mm}完\hspace{1mm}午饭\hspace{1mm}要\hspace{1mm}多\hspace{1mm}吃\hspace{1mm}点\hspace{1mm}水果 。\\
        Ours:& \textcolor{blue}{吃\hspace{1mm}完\hspace{1mm}午饭}\hspace{1mm}我\hspace{1mm}\textcolor{blue}{要\hspace{1mm}多\hspace{1mm}吃\hspace{1mm}点\hspace{1mm}水果 。} \\ \midrule

        Gloss:& 你们\hspace{1mm}吃\hspace{1mm}什么\hspace{1mm}我\hspace{1mm}请客 \\
        Vis. Token:& 看\colorbox{green!40}{你们}大家来\colorbox{green!40}{吃}饭\colorbox{green!40}{什么}他怕\colorbox{green!40}{我}\hspace{1mm}\colorbox{green!40}{请}十分钟\\
        \noalign{\vskip 0.3ex}\cdashline{1-2}\noalign{\vskip 0.7ex}
        Ref:& 你们\hspace{1mm}吃\hspace{1mm}点\hspace{1mm}什么,\hspace{1mm}我\hspace{1mm}请客 。 \\
        Ours:& \textcolor{blue}{你们\hspace{1mm}吃}饭\hspace{1mm}\textcolor{blue}{什么}?\hspace{1mm}\textcolor{blue}{我 请}\hspace{1mm}你们\hspace{1mm}吃饭 。\\ \midrule


        Gloss:& 超市\hspace{1mm}我\hspace{1mm}要\hspace{1mm}买\hspace{1mm}椅子\hspace{1mm}你\hspace{1mm}去 \\
        Vis. Token:& \colorbox{green!40}{超市} \colorbox{green!40}{买} 事 茶 怕 \colorbox{green!40}{我} \colorbox{green!40}{要} \colorbox{green!40}{我} \colorbox{green!40}{买} \colorbox{green!40}{椅子} 时候 什么 \colorbox{green!40}{你} 他 来 去 \colorbox{blue!40}{吗?} \\
        \noalign{\vskip 0.3ex}\cdashline{1-2}\noalign{\vskip 0.7ex}
        Ref:& 我\hspace{1mm}要\hspace{1mm}去\hspace{1mm}超市\hspace{1mm}买\hspace{1mm}椅子,\hspace{1mm}你\hspace{1mm}去\hspace{1mm}吗? \\
        Ours:& \textcolor{blue}{我\hspace{1mm}要\hspace{1mm}去\hspace{1mm}超市\hspace{1mm}买\hspace{1mm}椅子,\hspace{1mm}你\hspace{1mm}去\hspace{1mm}吗?} \\ \midrule 

        Gloss:& 我 \hspace{1mm}驾驶 \hspace{1mm}高铁 \hspace{1mm}站 \hspace{1mm}接 \hspace{1mm}儿子 \\
        Vis. Token:& \colorbox{green!40}{天} \colorbox{green!40}{气}大憾\colorbox{green!40}{冷}怕我\colorbox{blue!40}{我们}\hspace{1mm}\colorbox{blue!40}{去}下雨大学今天 \\
        \noalign{\vskip 0.3ex}\cdashline{1-2}\noalign{\vskip 0.7ex}
        Ref:& 天气 \hspace{1mm}太\hspace{1mm}冷\hspace{1mm}了,\hspace{1mm}我们\hspace{1mm}去\hspace{1mm}打 \hspace{1mm}篮球\hspace{1mm}吧 。 \\
        Ours:& \textcolor{blue}{天气\hspace{1mm}太\hspace{1mm}冷\hspace{1mm}了,\hspace{1mm}我们\hspace{1mm}去\hspace{1mm}打\hspace{1mm}篮球\hspace{1mm}吧 。} \\ \midrule

        Gloss:& 穿 \hspace{1mm}暖 \hspace{1mm}没有 \hspace{1mm}我 \hspace{1mm}想 \hspace{1mm}不 \hspace{1mm}买 \\
        Vis. Token:& \colorbox{blue!40}{衣服} 饱 热情 急 \colorbox{green!40}{没有} 怕 \colorbox{green!40}{我} \colorbox{green!40}{想} 觉得 想 \colorbox{green!40}{不} 不是 \colorbox{green!40}{买} 不是 \\
        \noalign{\vskip 0.3ex}\cdashline{1-2}\noalign{\vskip 0.7ex}
        Ref:& 这 \hspace{1mm}件 \hspace{1mm}衣服\hspace{1mm}不\hspace{1mm}保暖,\hspace{1mm}我\hspace{1mm}不\hspace{1mm}想\hspace{1mm}买 。 \\
        Ours:& \textcolor{blue}{这\hspace{1mm}件\hspace{1mm}衣服\hspace{1mm}不\hspace{1mm}保暖,\hspace{1mm}我\hspace{1mm}不\hspace{1mm}想\hspace{1mm}买 。} \\ \bottomrule
        
    \end{tabular}
}
\caption{Translation results on the CSL-Daily test set. Exact visual token matches within glosses are highlighted in \colorbox{green!40}{green}. Words highlighted in \colorbox{blue!40}{blue} are not present in the glosses but appear in the translation. Correctly translated 1-grams are shown in \textcolor{blue}{blue}.}
\label{tab:qaul_res_csl}
\end{table*}
\end{CJK*}

\begin{table*}[t]
\scriptsize
\centering
\resizebox{\linewidth}{!}{%
    \begin{tabular} {rl} \toprule 
        Image:& \includegraphics[width=0.6\linewidth]{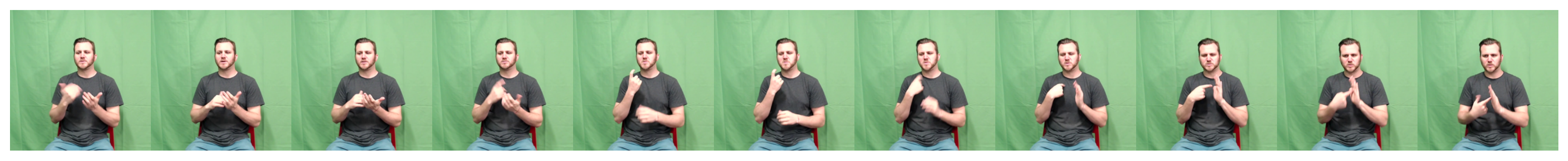} \\
        Vis. Token:& \colorbox{green!40}{AGAIN} SOMEONE \colorbox{green!40}{ONE} \colorbox{green!40}{SHOW} \\
        Ref:& again, one more time we'll show it for you. \\
        Ours:& \textcolor{blue}{again, one more time}. \\ \midrule 

        Image:& \includegraphics[width=0.6\linewidth]{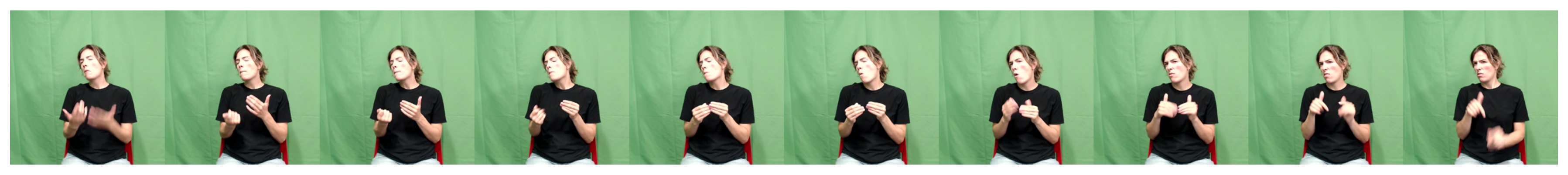} \\
        Vis. Token:& \colorbox{green!40}{LITTLE} \colorbox{green!40}{MORE} HOW \\
        Ref:& a little bit more then this maybe. \\
        Ours:& \textcolor{blue}{a little bit more} about it. \\ \midrule 

        Image:& \includegraphics[width=0.8\linewidth]{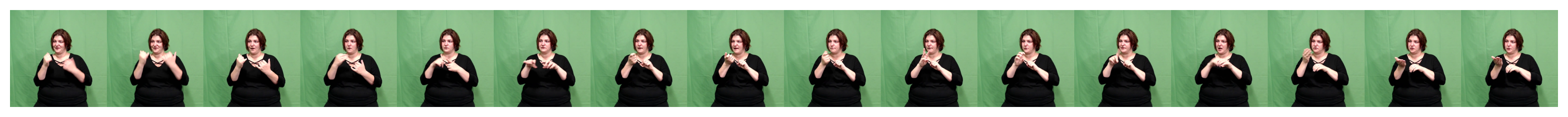} \\
        Vis. Token:& \colorbox{green!40}{NOW} GO TODAY TO TAKE \colorbox{green!40}{LITTLE} THREE SEVEN FOUR \colorbox{green!40}{WEED} \colorbox{green!40}{OUT} \colorbox{green!40}{LITTLE} \colorbox{green!40}{HERE} JUILLET VORSCHRIFTEN \\
        Ref:& and we're going to take a little weed out here. \\
        Ours:& \textcolor{blue}{now we're going to take a little} bit of the \textcolor{blue}{weed out here}. \\ \midrule 

        Image:& \includegraphics[width=0.9\linewidth]{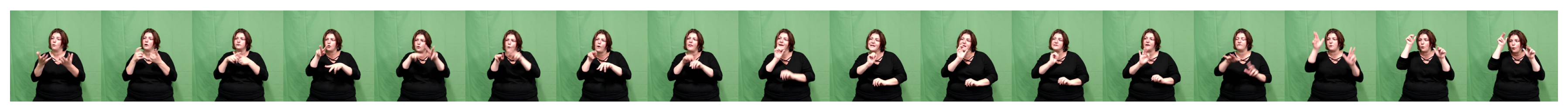} \\
        Vis. Token:& \colorbox{green!40}{WANT} TO REPEAT TWO \colorbox{green!40}{LOOK} ON YOURÄNG KISS AGE IS \colorbox{green!40}{YOUR} \colorbox{green!40}{HORSE} \\
        Ref:& you want to look at the age of your horse. \\
        Ours:& \textcolor{blue}{you want to} take a \textcolor{blue}{look at the age of your horse}. \\ \midrule 

        Image:& \includegraphics[width=0.6\linewidth]{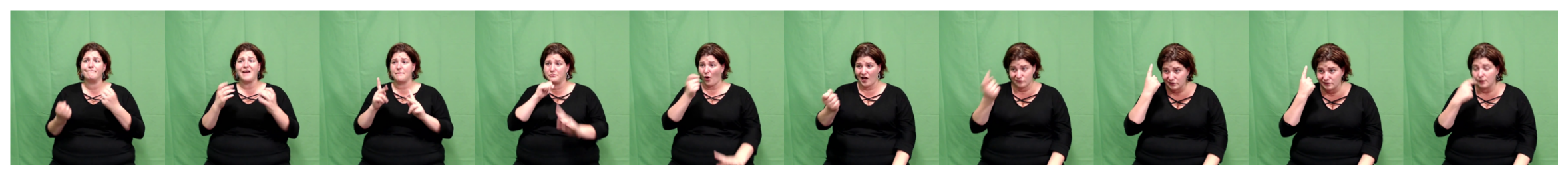} \\
        Vis. Token:& \colorbox{green!40}{MANY} \colorbox{green!40}{PEOPLE} \colorbox{green!40}{NOT} OTHER \colorbox{green!40}{UNDERSTAND} THOUGHT \\
        Ref:& many people don't understand. \\
        Ours:& \textcolor{blue}{many people don't understand} that. \\ \midrule 

        Image:& \includegraphics[width=0.8\linewidth]{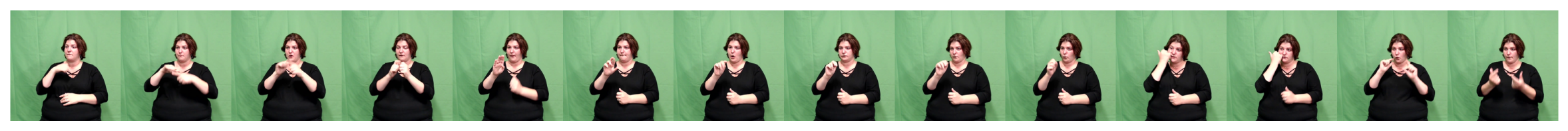} \\
        Vis. Token:& \colorbox{green!40}{I} \colorbox{green!40}{PRACTICE} WHEN \colorbox{green!40}{WITH} B FOAMERS CAST WAS SO OROU CAN KNOW IF OR \colorbox{green!40}{GROUP} \\
        Ref:& i practice with the barton oaks dental group. \\
        Ours:& \textcolor{blue}{i practice with the barton oaks} tennis team. \\ \midrule
        
        
                            
        Ref:& so, let's keep doing the same thing with the arms. \\
        Ours:& \textcolor{blue}{so, let's keep doing the same thing with the arms.} \\ \midrule
        
        Ref:& here, two, three, four, elbow and follow wherever you're going to go, like the knee to the groin and your elbow. \\
        Ours:& \textcolor{blue}{here, two, three, four,} \textcolor{blue}{follow} through where \textcolor{blue}{you're going to} want to squeegee, woo, woo, \textcolor{blue}{your elbow.} \\ \midrule 
        
        Ref:& my name is robert segundo and have fun. \\
        Ours:& \textcolor{blue}{my name is robert} todd \textcolor{blue}{and have fun.} \\ \midrule
        
        Ref:& watch our next segment to learn more about natural beauty products. \\
        Ours:& \textcolor{blue}{watch our next segment} and we'll talk a little bit \textcolor{blue}{more about natural beauty products.} \\ \midrule 

        Ref:& remember, be careful when doing your home remedies, and if you're not sure, check with your local professional. \\
        Ours:& \textcolor{blue}{remember} very carefully \textcolor{blue}{when doing your home remedies} \textcolor{blue}{if you} have a cell phone. \\ \midrule 

        Ref:& you can start to rotate your shoulders and start to get more comfortable with your feet by turning. \\
        Ours:& \textcolor{blue}{you can start} rotating \textcolor{blue}{your shoulders} and \textcolor{blue}{start} getting \textcolor{blue}{comfortable} \textcolor{blue}{with your} five \textcolor{blue}{by} rotating. \\ \midrule 

        Ref:& hi, i'm johanna krynytzky with hip expressions belly dance studio in st. petersburg, florida. \\
        Ours:& \textcolor{blue}{hi, i'm johanna krynytzky with hip expressions belly dance studio in st. petersburg, florida.} \\ \midrule 

        Ref:& i'm going to show you how to do some step-touch side foot work for belly dancing. \\
        Ours:& \textcolor{blue}{i'm going to show you} \textcolor{blue}{some step touch side} and medium rock \textcolor{blue}{for belly dancing.} \\ \bottomrule 
        
    \end{tabular}
}
\caption{Translation results on the How2Sign test set. Correctly translated 1-gram matches are highlighted in \textcolor{blue}{blue}. Exact visual token matches within the translation are highlighted in \colorbox{green!40}{green}.}
\label{tab:qaul_res_h2s}
\end{table*}

\end{document}